%% file: main.tex
\documentclass[10pt,journal,compsoc]{IEEEtran}
\usepackage{amsmath,amsfonts}
\usepackage{array}
\usepackage[caption=false,font=normalsize,labelfont=sf,textfont=sf]{subfig}
\usepackage{textcomp}
\usepackage{stfloats}
\usepackage{url}
\usepackage{verbatim}
\usepackage{graphicx}
\usepackage{cite}
\usepackage{booktabs}
\usepackage{tabularx}
\usepackage{multirow}
\usepackage{makecell}
\usepackage[table]{xcolor}
\usepackage{ragged2e}
\usepackage[ruled]{algorithm2e}

\SetAlFnt{\small}
\SetAlCapFnt{\small}
\SetAlCapNameFnt{\small}
\SetAlCapHSkip{0pt}

\usepackage{enumitem}
\setlist[enumerate]{leftmargin=5mm}
\setlist[itemize]{leftmargin=5mm}
\setlist[enumerate]{leftmargin=5mm}
\setlist[itemize]{leftmargin=5mm}

\usepackage[most]{tcolorbox}
\usepackage{colortbl}
\usepackage[pagebackref=true,breaklinks=true, colorlinks,bookmarks=false,linkcolor=blue,citecolor=blue]{hyperref}

\def\eg{\textit{e.g.}}
\newcommand{\revised}[1]{\textcolor{black}{#1}}

\begin{document}

\title{PanopticQuery: Unified Query-Time Reasoning for 4D Scenes}

\author{
    Ruilin~Tang,
    Yang~Zhou,
    Zhong~Ye,
    Wenxi~Liu,
    Yan~Huang,
    and Shengfeng~He,~\IEEEmembership{Senior Member,~IEEE}
    \IEEEcompsocitemizethanks{
        \IEEEcompsocthanksitem Ruilin Tang and Yang Zhou are with the School of Computer Science and Engineering, South China University of Technology, Guangzhou 510006, China; They are also with the School of Computing and Information Systems, Singapore Management University, Singapore 188065 (E-mail: t1344409@gmail.com; matrixgle19@gmail.com).
        \IEEEcompsocthanksitem Yan Huang is with the School of Computer Science and Engineering, South China University of Technology, Guangzhou 510006, China (E-mail: aihuangy@scut.edu.cn).
        \IEEEcompsocthanksitem Zhong Ye is with the School of Computer Science and Technology, Guangdong University of Technology, Guangzhou 510006, China (E-mail: zhongye0312@gmail.com).
        \IEEEcompsocthanksitem Wenxi Liu is with the College of Computer and Data Science, Fuzhou University, Fuzhou 350108, China (E-mail: wenxi.liu@hotmail.com).
        \IEEEcompsocthanksitem Shengfeng He is with the School of Computing and Information Systems, Singapore Management University, Singapore 188065 (E-mail: shengfenghe@smu.edu.sg).
    }
}

\markboth{IEEE Transactions on Pattern Analysis and Machine Intelligence}
{Tang \MakeLowercase{\textit{et al.}}: PanopticQuery: Unified Query-Time Reasoning for 4D Scenes}

\IEEEcompsoctitleabstractindextext{
\begin{abstract}
\justifying
Understanding dynamic 4D environments through natural language queries requires not only accurate scene reconstruction but also robust semantic grounding across space, time, and viewpoints. While recent methods using neural representations have advanced 4D reconstruction, they remain limited in contextual reasoning, especially for complex semantics such as interactions, temporal actions, and spatial relations. A key challenge lies in transforming noisy, view-dependent predictions into globally consistent 4D interpretations. We introduce PanopticQuery, a framework for unified query-time reasoning in 4D scenes. Our approach builds on 4D Gaussian Splatting for high-fidelity dynamic reconstruction and introduces a multi-view semantic consensus mechanism that grounds natural language queries by aggregating 2D semantic predictions across multiple views and time frames. This process filters inconsistent outputs, enforces geometric consistency, and lifts 2D semantics into structured 4D groundings via neural field optimization. To support evaluation, we present Panoptic-L4D, a new benchmark for language-based querying in dynamic scenes. Experiments demonstrate that PanopticQuery sets a new state of the art on complex language queries, effectively handling attributes, actions, spatial relationships, and multi-object interactions. A video demonstration is available in the supplementary materials.
\end{abstract}

\begin{IEEEkeywords}
4D Scene Understanding, Multimodal Learning, Language Grounding.
\end{IEEEkeywords}
}

\maketitle
\IEEEdisplaynotcompsoctitleabstractindextext
\IEEEpeerreviewmaketitle

\begin{figure*}
\centering
  \includegraphics[width=\linewidth]{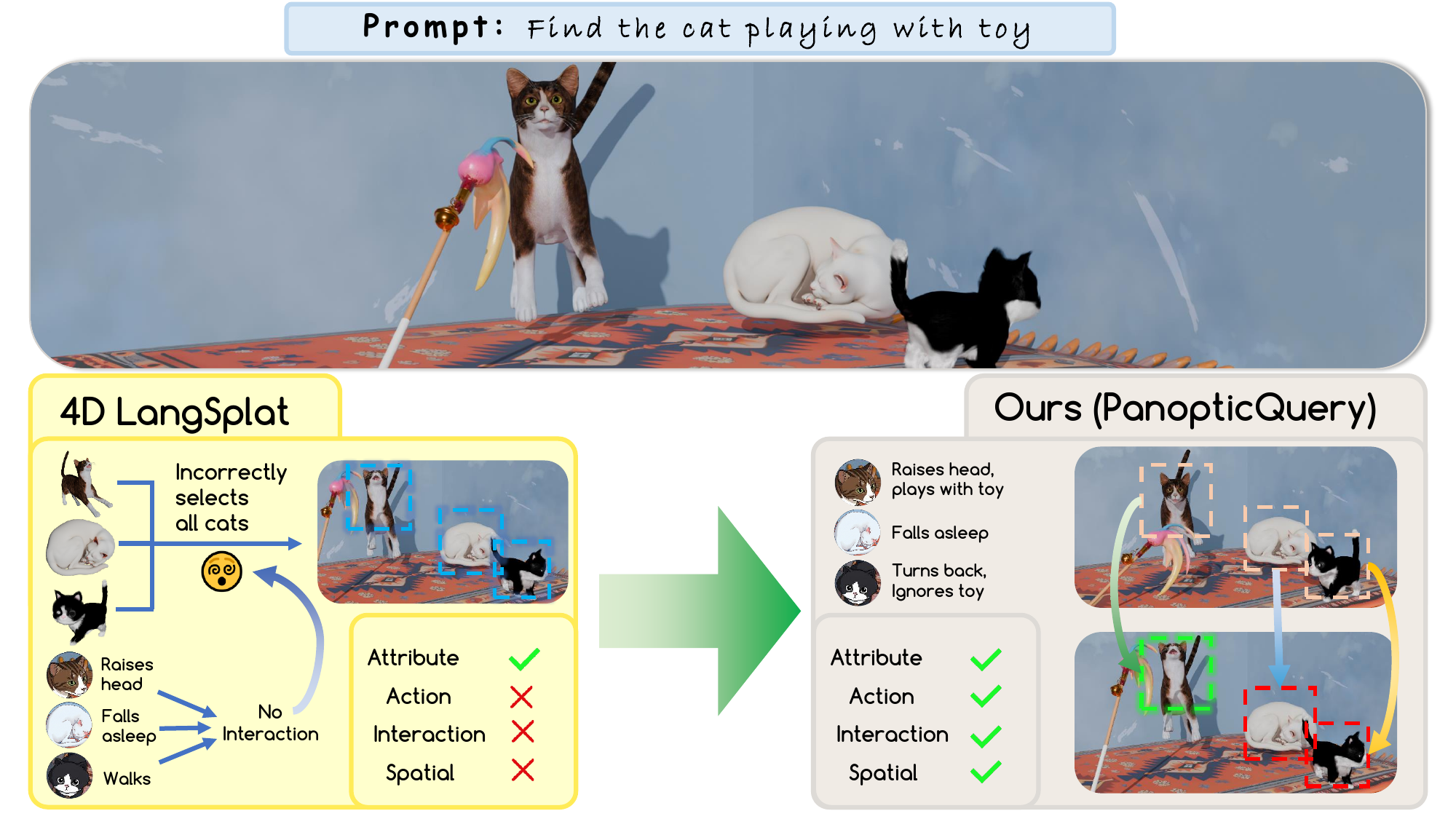}
  \vspace{-8mm}
  \caption{
    While state-of-the-art embedding-based methods, such as 4D LangSplat, perform well on static attribute queries, they struggle with actions and interactions, often collapsing multiple instances of the same category into a single response, for example selecting all cats (left). PanopticQuery introduces query-time reasoning that leverages spatio-temporal context to disambiguate entities based on their interaction patterns, enabling precise grounding of the target object, such as the cat actively playing with the toy, where prior methods fail (right). This example is from our newly proposed Panoptic-L4D dataset.}
  \label{fig:teaser}
\end{figure*}

\input{sec/1_intro}
\input{sec/2_related}
\input{sec/3_dataset}
\input{sec/4_method}
\input{sec/5_experiment}
\input{sec/6_conclusion}

\bibliographystyle{IEEEtran}
\bibliography{ref}

\begin{IEEEbiography}[{\includegraphics[width=1in,height=1.25in,clip,keepaspectratio]{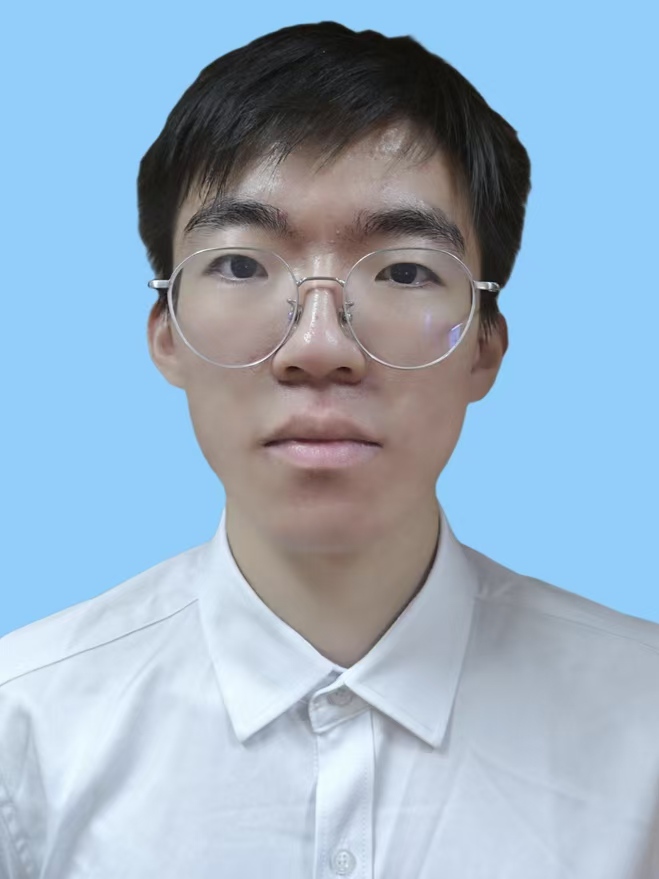}}]{Ruilin Tang} is currently pursuing the B.S. degree with the School of Computer Science and Engineering, South China University of Technology. His research interests include computer vision, 4D scene understanding, and neural rendering.
\end{IEEEbiography}

\begin{IEEEbiography}[{\includegraphics[width=1in,height=1.25in,clip,keepaspectratio]{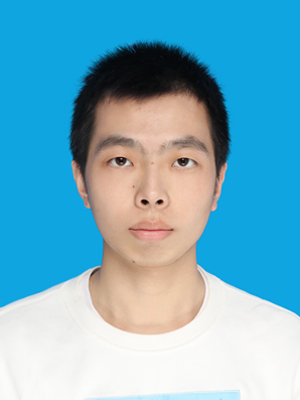}}]{Yang Zhou} received a B.Sc. degree from the School of Computer Science and Engineering, South China University of Technology, in 2021, where he is currently pursuing a Ph.D. degree. His research interests include computer vision, image processing, and deep learning.
\end{IEEEbiography}

\begin{IEEEbiography}[{\includegraphics[width=1in,height=1.25in,clip,keepaspectratio]{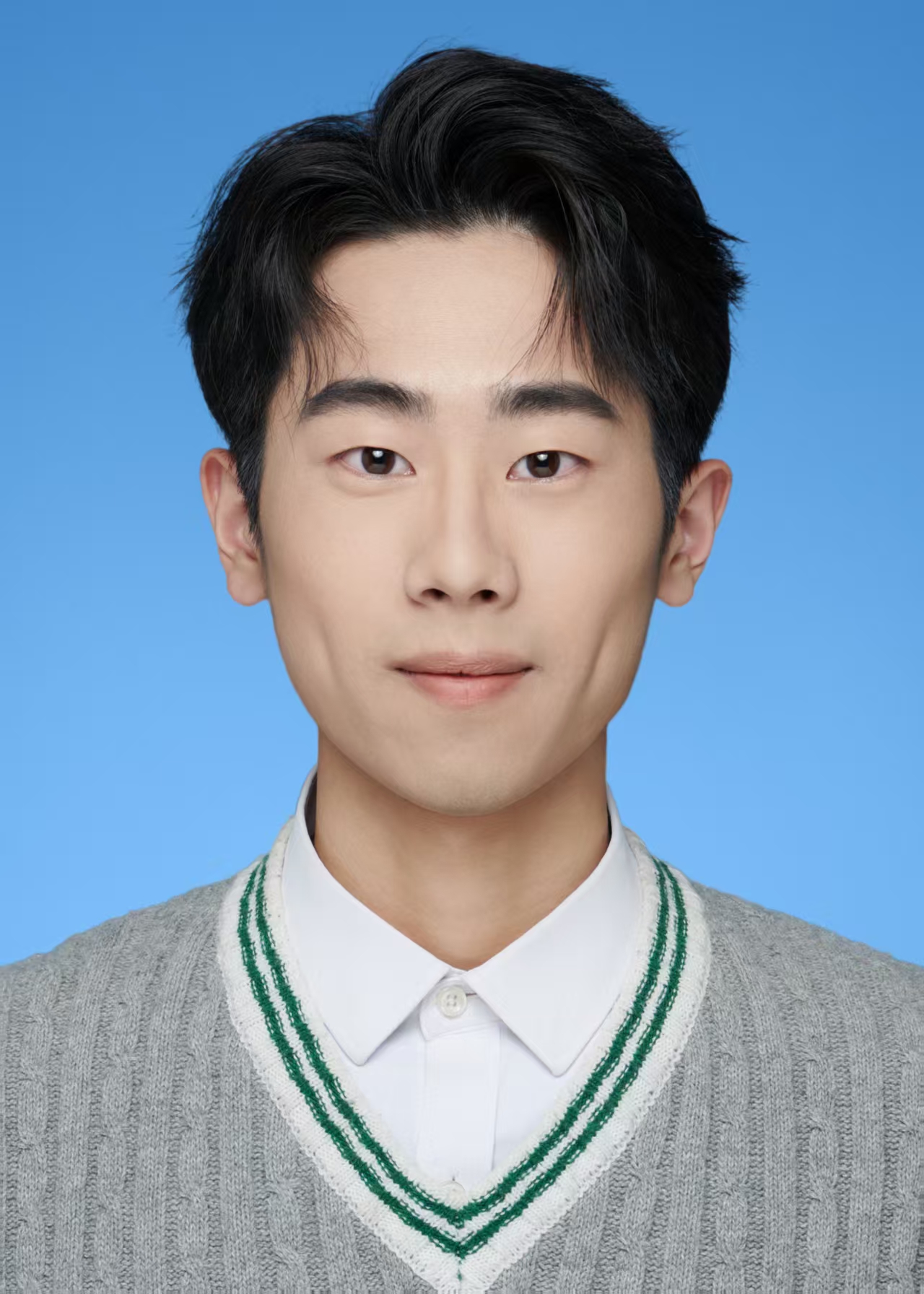}}]{Zhong Ye} is a senior undergraduate student at Guangdong University of Technology. His research interests include  dynamic adaptation, memory in intelligent systems and computer vision.
\end{IEEEbiography}

\begin{IEEEbiography}[{\includegraphics[width=1in,height=1.25in,clip,keepaspectratio]{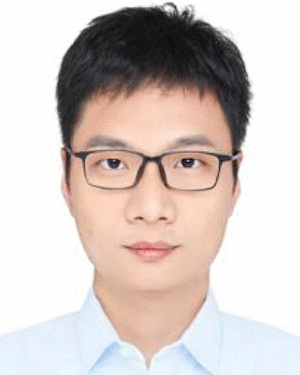}}]{Wenxi Liu} received the PhD degree from the City University of Hong Kong. He is currently a professor with the College of Computer and Data Science, Fuzhou University. His research interests include computer vision, robot vision, and image processing.
\end{IEEEbiography}

\begin{IEEEbiography}[{\includegraphics[width=1in,height=1.25in,clip,keepaspectratio]{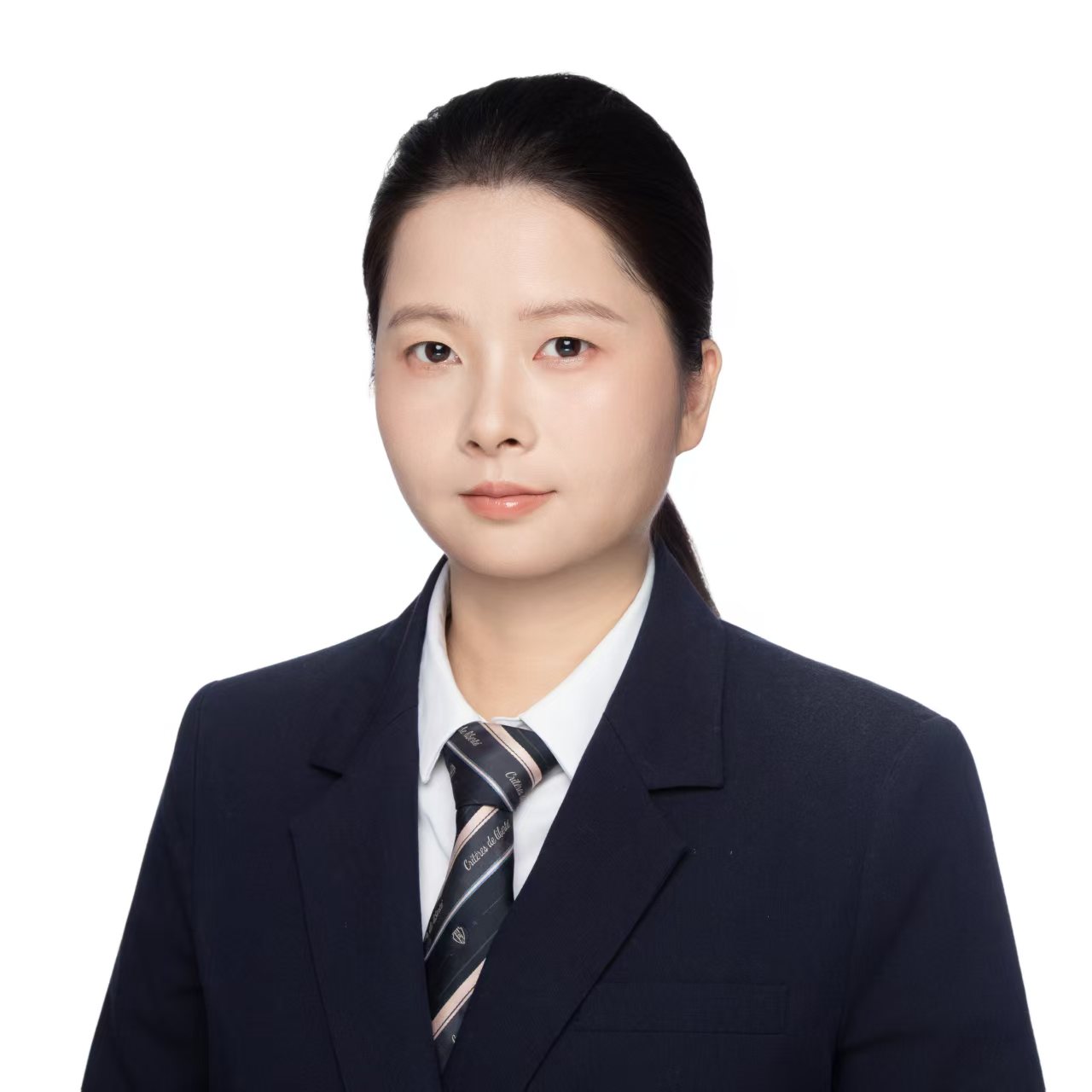}}]{Yan Huang} is currently an Associate Professor and Ph.D. supervisor with the School of Computer Science and Engineering, South China University of Technology. She received the B.S. degree from Hunan University, in 2013, and the Ph.D. degree from South China University of Technology, in 2018. Her research interests include image processing, pattern recognition, and artificial intelligence.
\end{IEEEbiography}

\begin{IEEEbiography}[{\includegraphics[width=1in,height=1.25in,clip,keepaspectratio]{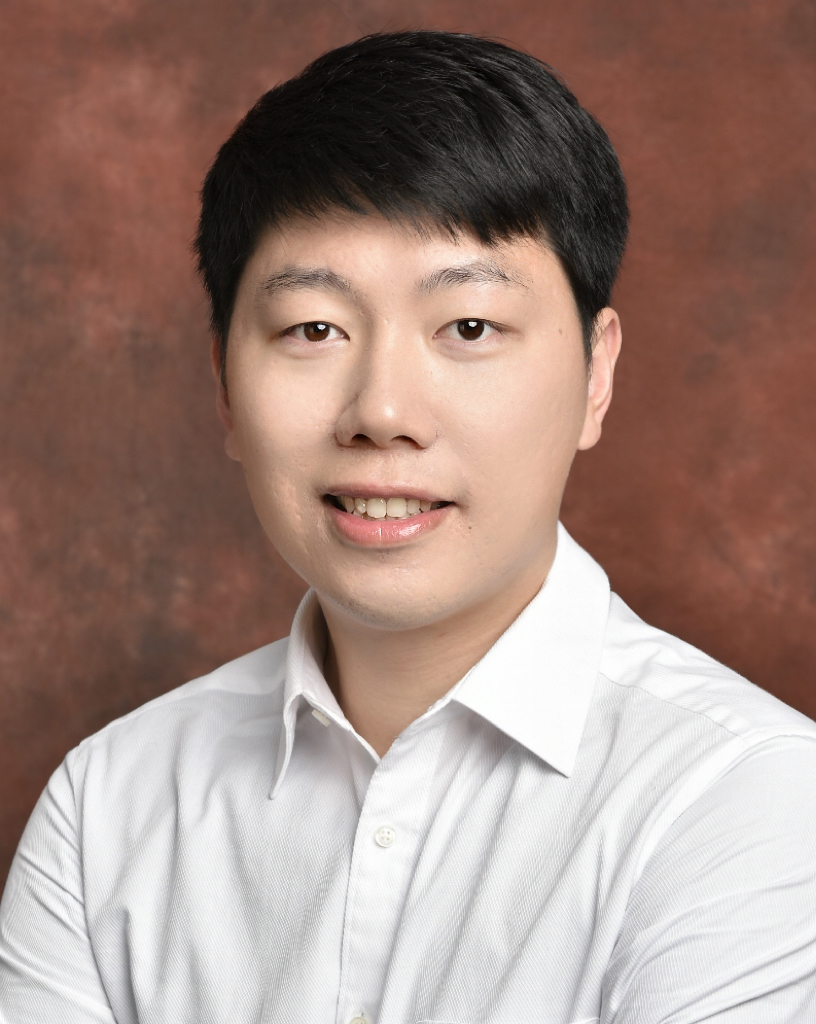}}]{Shengfeng He (Senior Member, IEEE)} is an associate professor in the School of Computing and Information Systems at Singapore Management University. He earned his B.Sc. and M.Sc. from Macau University of Science and Technology (2009, 2011) and a Ph.D. from City University of Hong Kong (2015). His research focuses on computer vision and generative models. He has received awards including the Google Research Award, PerCom 2024 Best Paper Award, and the Lee Kong Chian Fellowship. He is a senior IEEE member and a distinguished CCF member. He serves as lead guest editor for IJCV and associate editor for IEEE TPAMI, IEEE TNNLS, IEEE TCSVT, Visual Intelligence, and Neurocomputing. He is a senior area chair for NeurIPS, an area chair for CVPR, NeurIPS, ICLR, ICML, AAAI, IJCAI, BMVC, and the Conference Chair of Pacific Graphics 2026.
\end{IEEEbiography}

\end{document}

%% file: sec/1_intro.tex
\section{Introduction}
\label{sec:intro}

\IEEEPARstart{T}he ultimate goal of scene understanding is to develop AI systems that perceive and reason about the world in a human-like manner, not as collections of static objects with fixed labels, but as dynamic environments where entities interact, events unfold over time, and meaning emerges from context. Achieving this level of understanding in 4D scenes, where both space and time are continuous, is critical for applications in robotics, augmented reality, and human-AI interaction. Intelligent agents must be capable of answering open-ended, compositional queries such as ``Where is the red cup?'' or “What did the person just do with the book?”, grounding their responses in the scene’s structure and temporal dynamics.

Recent advances in neural representations, including neural radiance fields~\cite{mildenhall2021nerf, park2021hypernerf, park2021nerfies} and 3D Gaussian Splatting~\cite{kerbl20233d}, have enabled high-fidelity reconstruction of dynamic scenes~\cite{wu20244d, yang2024deformable}. Building on these developments, methods like 4D LangSplat~\cite{li20254d} attempt to integrate language by embedding precomputed semantic features into the 4D representation, associating each point with a static descriptor. While effective for basic attribute-based queries, such as “the red cup”, this embed-first, query-later approach encodes semantics rigidly into the scene, making it difficult to reason about novel, relational, or temporal concepts.

We argue that true scene understanding is contextual and compositional. For example, the meaning of “the mouse being moved by the human” is not attributable to the mouse or the human in isolation, but emerges from their spatial and temporal relationship, as well as the action implied. As shown on the left of Fig.~\ref{fig:teaser}, static feature fields often fail to distinguish such interactions. They recognize objects but miss the narrative connecting them, leading to incorrect or ambiguous grounding, such as selecting all similar objects regardless of behavior.

We propose a new paradigm for 4D scene understanding: decoupling the scene representation from semantic reasoning. Our method, \textit{PanopticQuery}, maintains a purely geometric and photometric 4D representation using 4D Gaussian Splatting~\cite{wu20244d}, and performs semantic reasoning dynamically at query time. This allows the scene to remain conceptually open, while enabling flexible interpretation depending on the query.

At inference time, given a language query and a time range, our system renders multi-view RGB and depth video clips from the 4D scene. These clips are processed by a multi-view semantic grounder that generates 2D segmentation masks conditioned on the prompt. Since these masks are often noisy and inconsistent across views, we introduce a \textit{Multi-View Semantic Consensus} module to fuse them into coherent 4D predictions. This module performs geometric voting across views and frames to enforce spatial consistency and suppress ambiguity. Finally, we fit a lightweight neural field~\cite{ye2024gaussian, ji2024segment} to the fused predictions, producing a temporally consistent and geometry-aware 4D grounding.

As shown on the right of Fig.~\ref{fig:teaser}, this query-time reasoning framework enables PanopticQuery to resolve fine-grained semantics, such as identifying “the cat playing with toy” while ignoring others that are asleep or disinterested. In contrast, embedding-based methods like 4D LangSplat conflate all instances of the same category, lacking the ability to distinguish them through behavior or interaction.

To evaluate the full spectrum of 4D scene understanding, we define four types of language queries:
\begin{itemize}
\item \textbf{Attribute queries} (What is it?)
\item \textbf{Action queries} (What is it doing?)
\item \textbf{Spatial queries} (Where is it relative to something else?)
\item \textbf{Interaction queries} (How do multiple objects relate over time?)
\end{itemize}

These dimensions are underexplored in current 4D benchmarks~\cite{pumarola2021d, park2021hypernerf, li2022neural}, which focus primarily on static object reconstruction or basic attributes. To support comprehensive evaluation, we introduce the \textit{Panoptic Language for 4D} (Panoptic-L4D) dataset, the first benchmark to target 4D dynamic scene understanding through open-ended natural language queries. Panoptic-L4D features richly annotated dynamic scenes designed to probe contextual, relational, and temporal reasoning across a wide range of semantic tasks. 

Our contributions are as follows:
\begin{itemize}
\item We propose \textit{PanopticQuery}, a novel framework that decouples 4D scene representation from language reasoning. By performing spatio-temporal interpretation at query time, our approach enables a generalizable understanding of dynamic scenes.
\item We present the first method that robustly supports complex language queries across multiple categories, including attributes, actions, spatial relationships, and object interactions.
\item We introduce a Multi-View Semantic Consensus algorithm that aggregates noisy, view-dependent 2D outputs into a coherent 4D spatio-temporal grounding through geometric consistency and neural field learning.
\item We construct \textit{Panoptic-L4D}, the first benchmark designed to evaluate 4D scene understanding via contextual, relational, and temporal language queries in dynamic environments.
\end{itemize}

%% file: sec/2_related.tex
\section{Related Works}
\label{sec:related}

Our work on open-vocabulary 4D scene querying builds upon and diverges from several key areas of research: dynamic 3D scene representation, language-grounding in 3D, and the integration of large foundation models.

\subsection{4D Scene Representation}
Representing 4D scenes is a foundational task in computer vision and graphics. Early approaches used deformation fields~\cite{park2021nerfies, pumarola2021d, song2023nerfplayer} or time-conditioned neural radiance fields~\cite{xian2021space, cao2023hexplane, fridovich2023k} to model scene changes. Recently, 3D Gaussian Splatting (3DGS)~\cite{kerbl20233d} has emerged as a high-quality, efficient alternative to neural fields. Its explicit representation naturally extends to dynamic scenes~\cite{wu20244d, yang2024deformable, li2024spacetime, Song_2025_ICCV, liu2025modgs}. Methods like 4D Gaussian Splatting (4DGS)~\cite{wu20244d} modeling Gaussians whose properties (position, rotation, scale) change over time. Our work is built upon a 4DGS backbone, leveraging its ability for fast, high-fidelity rendering of dynamic scenes, which is crucial for our query-time video rendering pipeline.

\subsection{Language-Grounded 3D and 4D Understanding}
A significant research direction focuses on grounding natural language in 3D scenes. A dominant paradigm, exemplified by LERF~\cite{kerr2023lerf} and its 3DGS-based successors like LangSplat~\cite{qin2024langsplat}, is to distill features from models like CLIP into the 3D representation.
LangSplatV2~\cite{li2025langsplatv2} further extend it to high-dimensional semantics with faster speed.
Each point or Gaussian is assigned a static semantic embedding, allowing for open-vocabulary queries based on similarity search. ReferSplat~\cite{he2025refersplat} proposes a spatially aware framework that explicitly models 3D Gaussians with natural language expressions to enhance the robustness of referring segmentation against visual obstructions. Beyond feature distillation, recent approaches like Seeground~\cite{li2025seeground} and FreeQ-Graph~\cite{zhan2025freeq} explicitly leverage the generative and reasoning capabilities of 2D Vision-Language Models (VLMs) to address 3D scene understanding and visual grounding tasks. This approach has been extended to dynamic 4D scenes in works such as DGD~\cite{labe2024dgd}, 4-LEGS~\cite{fiebelman20254}, Feature4X~\cite{zhou2025feature4x} and 4D LangSplat~\cite{li20254d}. While effective for attribute-based queries (``red cube''), a core limitation of these methods is their reliance on pre-computed, static features. These frozen embeddings cannot capture the dynamic, relational semantics of actions or interactions that emerge from the temporal context, such as ``the person who is waving.''
Our method fundamentally departs from this ``embed-first'' paradigm. Instead of baking semantics into the Gaussians, we maintain a purely geometric 4DGS representation and perform all semantic reasoning at query time.

\subsection{Foundation Models for Scene Understanding}
The rise of powerful foundation models has created new opportunities for scene understanding. The Segment Anything Model (SAM) ~\cite{kirillov2023segment} provides robust, category-agnostic segmentation capabilities. Subsequently, Multimodal Large Language Models (MLLMs) like GPT-4V~\cite{2023GPT4VisionSC} and Qwen-VL series~\cite{Qwen-VL, Qwen2-VL, Qwen2.5-VL, Qwen3-VL},  have demonstrated remarkable abilities for complex visual reasoning and following instructions. By incorporating temporal modeling, SAMWISE~\cite{cuttano2025samwise} further extends the capabilities of SAM2~\cite{ravi2024sam} to video understanding with natural language guidance.
We use such an MLLM as our basic semantic grounder to understand complex concepts from open-ended language queries and bridge the gap between reasoning and geometry, finally enable dynamic, relational, and holistic scene understanding beyond the reach of prior paradigms.

%% file: sec/3_dataset.tex
\section{Panoptic-L4D Dataset}
\label{sec:dataset}

A primary limitation of existing benchmarks for dynamic scene understanding~\cite{li2022neural, park2021hypernerf, pumarola2021d} is their inability to evaluate fine-grained semantic reasoning. Queries in these datasets often distinguish between semantically distant categories (\eg, 
``person'' vs. ``dog'') or rely on simple visual attributes, failing to test a model's grasp of subtle, context-dependent differences. Such settings are insufficient for benchmarking the complex relational and temporal understanding that our work targets.

To address this critical shortcoming, we introduce the Panoptic Language for 4D (Panoptic-L4D) dataset. Panoptic-L4D is designed around a core philosophy of resolvable ambiguity: scenes are intentionally constructed to include multiple visually similar objects where disambiguation requires reasoning about one of the four key dimensions we target: attributes, actions, spatial relationships, or interactions. For example, a model must distinguish between a cat that is \textit{sleeping} and another that is \textit{walking}, or luggage \textit{on the floor} versus luggage \textit{being held by a person}. This forces a model to move beyond simple object recognition and engage in deeper spatio-temporal analysis.

\begin{figure}[t]
    \centering
    \includegraphics[width=\linewidth]{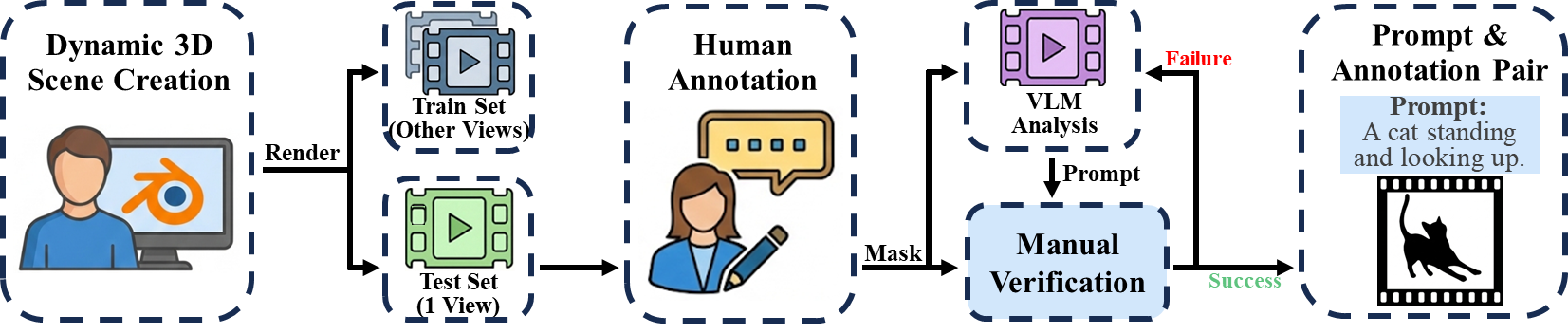}
    \vspace{-6mm}
    \caption{The Panoptic-L4D Construction Pipeline. Our two-phase process ensures high-quality, unambiguous data. An LLM agent proposes diverse scenarios, which are then manually selected and modeled in Blender to create realistic 4D scenes. The candidate (prompt, mask) pairs consist of human-annotated masks and VLM-generated prompts. These pairs then undergo an annotation-verification loop, where human annotators must unanimously agree on the prompt's referent, eliminating ambiguity before inclusion in the final benchmark.}
    \label{fig:dataset_pipeline}
    \vspace{-3mm}
\end{figure}

\begin{figure*}[t]
    \centering
    \includegraphics[width=\linewidth]{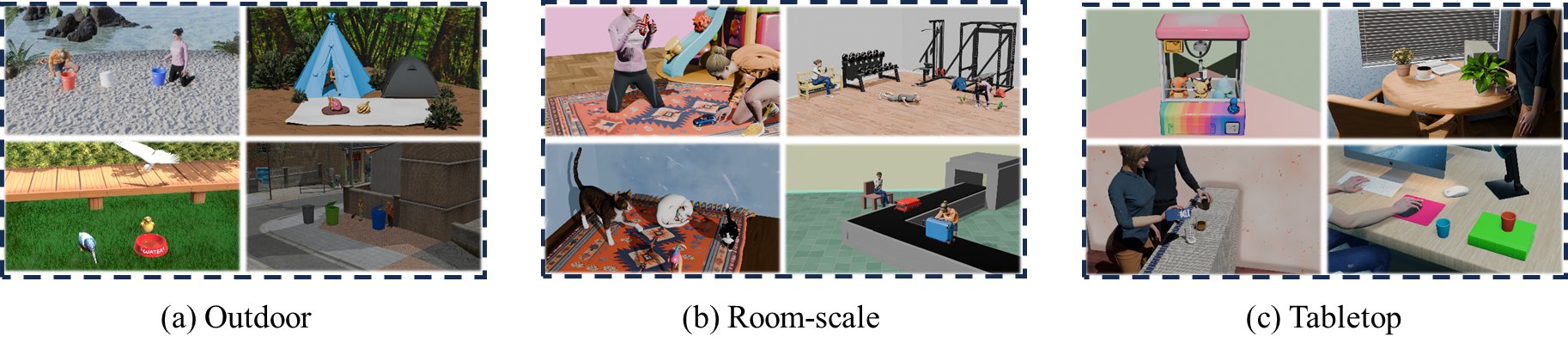}\vspace{-5mm}
    \caption{Examples from the Panoptic-L4D Benchmark. Our dataset spans diverse environments, including outdoor scenes, room-scale interactions, and complex tabletop arrangements. The queries challenge models to understand fine-grained actions, interactions, and spatial relations.}
    \label{fig:dataset_examples}
\end{figure*}

\subsection{Dataset Construction}
\label{sec:construction}

The construction of Panoptic-L4D follows the two-phase pipeline illustrated in Fig.~\ref{fig:dataset_pipeline}, ensuring that our scenes are both visually rich and linguistically precise.

\subsubsection{Phase 1: Scene Creation and Rendering.}
To achieve diversity, we employ a hybrid "agent-driven, human-refined" generation process. An LLM-based agent is prompted to generate high-level descriptions for dynamic scenarios spanning three scales: large outdoor environments, room-scale indoor facilities, and tabletop settings. From these proposals, we manually select 21 representative scenarios and construct them in Blender with expert refinement. Each dynamic 3D scene is rendered into multi-view videos. Following the protocol of 4D LangSplat~\cite{li20254d}, we randomly hold out one camera viewpoint for testing and use the remaining views for training the 4D scene representation.

\subsubsection{Phase 2: VLM Annotation and Manual Verification.}
To generate complex queries and their corresponding ground truth mask, we design an annotation-verification loop. First, for a given 4D scene, a powerful Vision-Language Model (VLM) analyzes the rendered videos with human pre-annotated objects and generates a rich set of candidate language queries aligned with our four target categories: Attributes, Actions, Spatial Relationships, and Interactions.

A crucial step is to ensure that each query has a single, unambiguous referent. To this end, we introduce a manual verification process. As shown in the right of Fig.~\ref{fig:dataset_pipeline}, 5 human annotators are tasked with identifying the target object(s) based only on the VLM-generated prompt.
\begin{itemize}
    \item Validation Criterion: A (prompt, mask) pair is considered valid only if all the human annotators successfully identify the correct object, ensuring unanimous consensus with ground truths.
    \item Refinement Loop: If any ambiguity arises (\eg, disagreement among annotators or identification of the wrong object), the (prompt, mask) pair is rejected and sent back for refinement or discarded entirely.
\end{itemize}
This strict verification cycle guarantees that Panoptic-L4D provides high-quality, discriminative ground truth for evaluating fine-grained 4D scene understanding.

\input{table/dataset_comparison}

\subsection{Dataset Analysis}

\subsubsection{Statistical Overview.}
Panoptic-L4D consists of 21 unique synthetic dynamic scenes and 4 real-world scenes. The synthetic categorized into 8 outdoor, 6 room-scale, and 7 tabletop environments (see Fig.~\ref{fig:dataset_examples} for examples). From these scenes, we have generated 289 high-quality, verified language queries corresponding to 108 distinct object instances. The number of queries are meticulously balanced to facilitate comprehensive evaluation across our four defined capabilities: Attribute (100), Action (61), Spatial (72), and Interaction (56).

To increase the diversity of the dataset, we additionally annotated four real-world scenes (birthday, train, theater, and painter, sourced from the InterDigital light field video dataset~\cite{sabater2017dataset}).
These four scenes serve as validation for our complex queries in the real world. 
The number of queries across our four defined capabilities: Attribute (15), Action (4), Spatial (13), and Interaction (4).

\subsubsection{Comparison with Prior Benchmarks.}
Tab.~\ref{tab:dataset_comparison} situates Panoptic-L4D relative to existing language-grounded datasets. While prior work has been foundational, our benchmark introduces a significant leap in both query volume and semantic complexity. For example, where 4D LangSplat provides 30 prompts for the Neu3D dataset, Panoptic-L4D offers nearly a 11$\times$ increase in linguistic annotations with far greater diversity. Critically, while previous benchmarks are saturated with simple attribute queries, Panoptic-L4D dedicates nearly 65\% of its queries to challenging action, spatial, and interaction-based reasoning tasks.

\subsubsection{Linguistic Complexity.}
\begin{figure}[t]
    \centering
    \subfloat[Object and scene nouns\label{fig:word_clouds_a}]{%
        \includegraphics[width=0.46\linewidth]{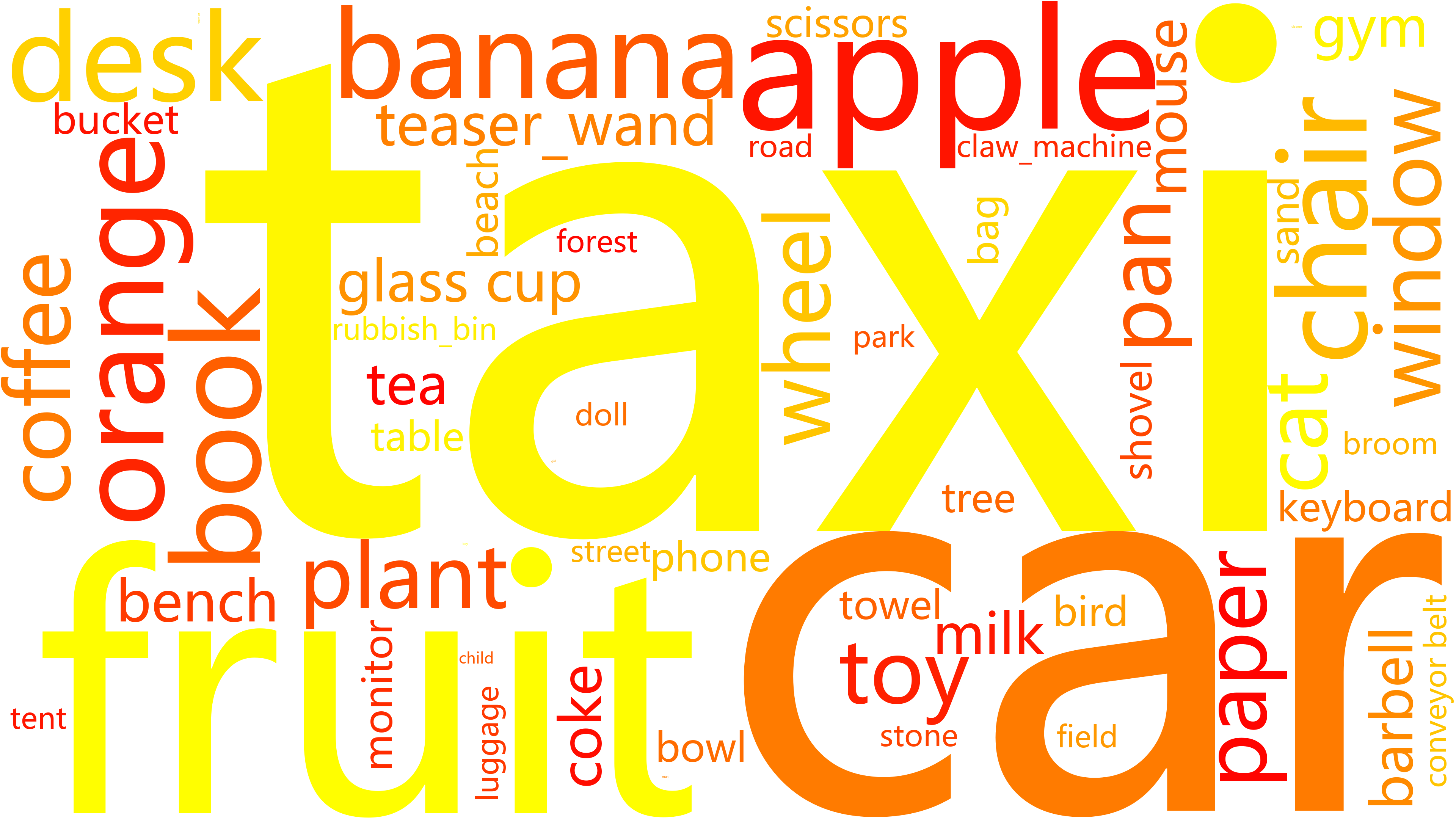}
    }
    \hfill
    \subfloat[Attr. and action modifiers\label{fig:word_clouds_b}]{%
        \includegraphics[width=0.46\linewidth]{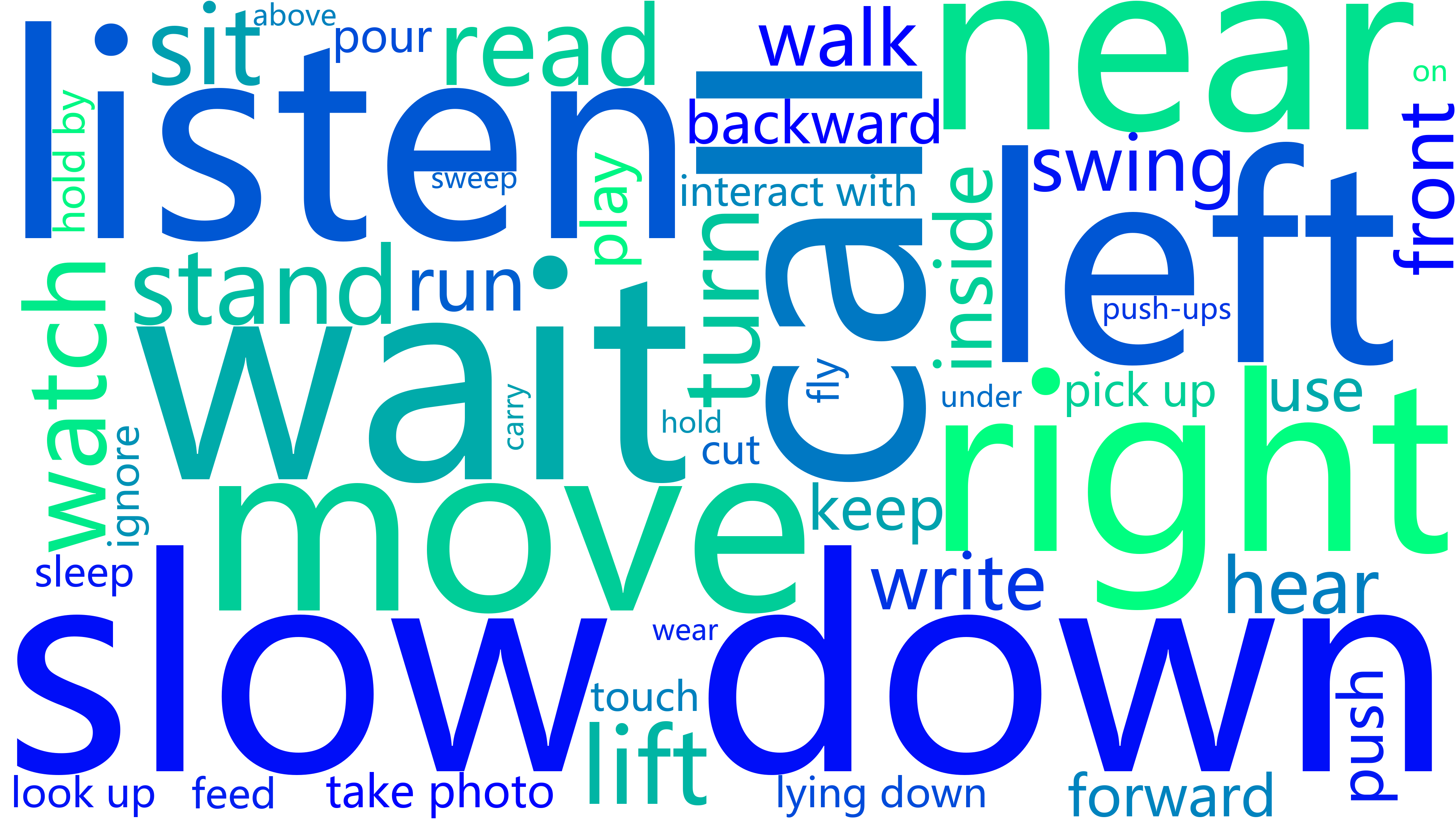}
    }
    \vspace{-2mm}
  \caption{Linguistic Diversity in Panoptic-L4D. Word clouds visualize the vocabulary of our benchmark. (a) Nouns describing common objects and scene types. (b) The emphasis on verbs (\eg, walking, holding, sleeping) and prepositions (\eg, on, inside, behind) highlights the dataset's focus on relational and dynamic reasoning, moving beyond simple attribute matching.}
  \label{fig:word_clouds}
\end{figure}

The linguistic richness of Panoptic-L4D is highlighted by the word clouds in Fig.~\ref{fig:word_clouds}. Unlike attribute-centric datasets, which are dominated by nouns and adjectives, Panoptic-L4D features a prominent vocabulary of verbs, prepositions, and relational phrases. This reflects the design goal of the benchmark: to compel models to move beyond matching static object properties and instead interpret dynamic events and spatial configurations. Queries require distinguishing "the person \textit{making a phone call}" from "the person \textit{feeding a cat}", a task that is impossible without a holistic understanding of actions and relationships unfolding over time. This shift from a noun-heavy to a verb- and preposition-rich query set marks a crucial step towards evaluating true 4D scene understanding.

%% file: table/dataset_comparison.tex
\begin{table}[t]
    \centering
    \caption{\textbf{Comparison with existing 4D datasets.} 
    Ours is the only dataset covering \textit{full-scale} scenarios (Tabletop, Room, Outdoor) with 
    \textit{ambiguous (action / spatial relation / interaction)} annotations. 
    ($^*$: Prompts and annotations adapted from 4D LangSplat~\cite{li20254d}.)}
    \vspace{-3mm}
    \label{tab:dataset_comparison}

    \setlength{\tabcolsep}{1pt}

    \resizebox{\linewidth}{!}{
    \rowcolors{2}{gray!20}{white}
    \begin{tabular}{lccccccc}
        \toprule
        \hline
        \textbf{Dataset} & \textbf{Type} & \textbf{Scale} & \textbf{Scenes} & 
        \textbf{Setup} & \textbf{Prompts} & \textbf{GT-Objs} & \textbf{Ambig.} \\
        \midrule
        D-NeRF        & Syn.  & Object   & 8  & Mono.            & --        & --        & $\times$ \\
        HyperNeRF     & Real  & Object   & 17 & Mono.            & 30$^*$    & 30$^*$    & $\times$ \\
        Neu3D         & Real  & Tabletop & 6  & Multi-view / Fwd.& 30$^*$    & 30$^*$    & $\times$ \\
        \midrule
        \rowcolor{gray!20}
        Ours & S/R  & \textbf{T / R / O} & \textbf{25} &
        Multi-view / Fwd. & \textbf{325} & \textbf{124} & \boldmath$\checkmark$ \\
        \hline
        \bottomrule
    \end{tabular}
    }

    \vspace{2pt}
    \begin{flushleft}
        \footnotesize
        \textit{Scale Legend:} \textbf{T}: Tabletop,\;
        \textbf{R}: Room,\;
        \textbf{O}: Outdoor. \\
        \textit{Setup Legend:} \textbf{Mono.}: Monocular,\;
        \textbf{Fwd.}: Forward-facing (180$^\circ$).
    \end{flushleft}
    \vspace{-3mm}
\end{table}

%% file: sec/4_method.tex
\begin{figure*}[t] 
    \centering 
    \includegraphics[width=1.0\textwidth]{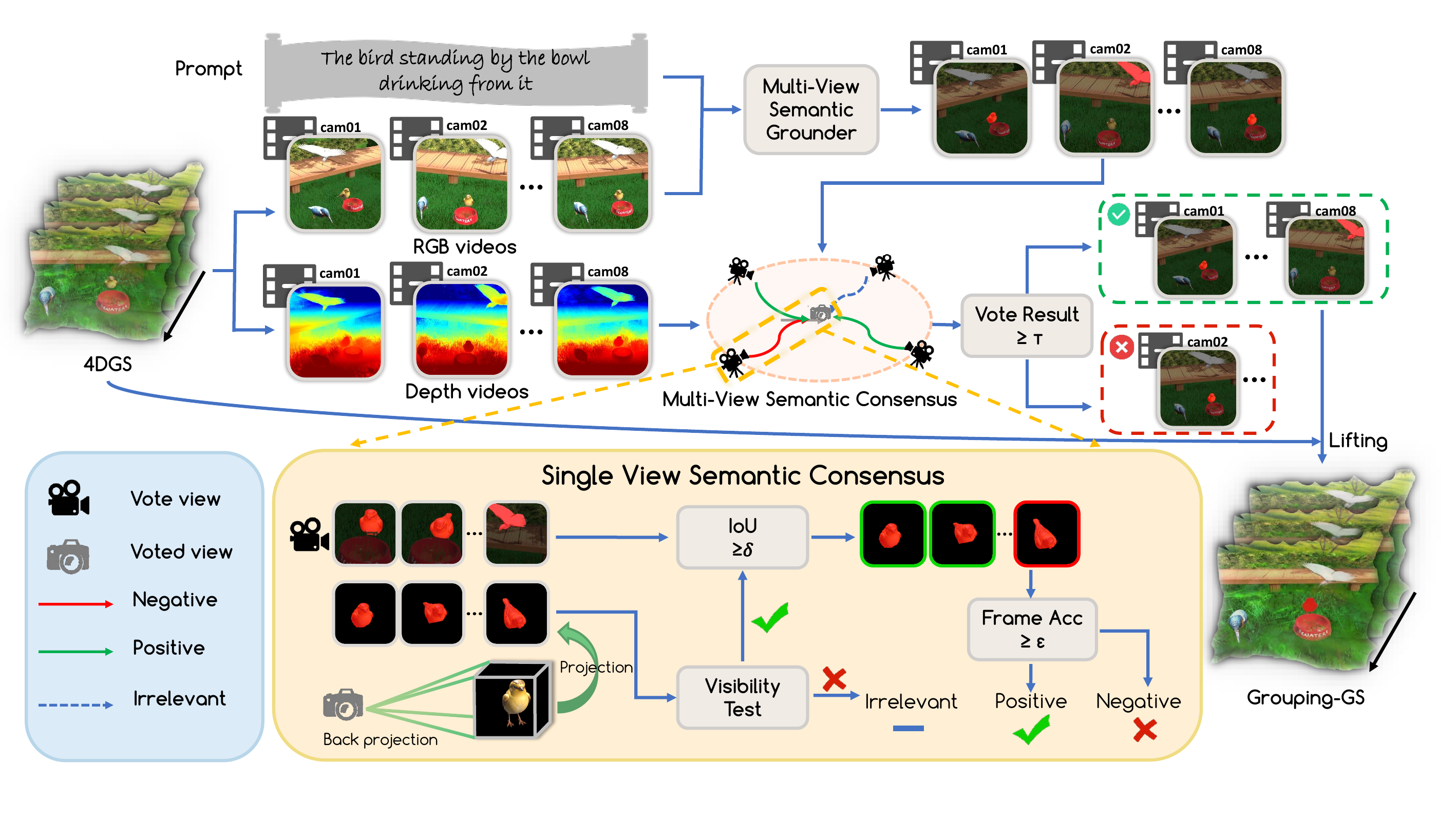} \vspace{-12mm}
    \caption{Overview of PanopticQuery. We render multi-view RGB/depth videos from an initial 4DGS and obtain prompt-conditioned masks with a multi-view semantic grounder. Our multi-view semantic consensus validates masks by back-projecting them to 3D and re-projecting to other views, discarding irrelevant views and retaining consistent ones (e.g., $\mathrm{IoU}\!>\!\delta$ and frame accuracy $>\!\epsilon$); masks supported by enough votes ($\ge\tau$) are lifted to the 4DGS for grouping-based embedding, enabling spatio-temporal querying.}
    \label{fig:pipeline} 
\end{figure*}

\section{PanopticQuery}
\label{sec:method}
\subsection{Preliminaries: 4D Gaussian Splatting}
Our framework leverages 4D Gaussian Splatting (4DGS)~\cite{wu20244d} as the underlying representation for dynamic scenes. In 4DGS, a dynamic scene is parameterized by a collection of deformable 3D Gaussians, $G = \{g_i\}_{i=1}^N$, where parameters evolve over time $t$. Each Gaussian $g_i$ is defined by its time-variant center $\mu_i(t)$, covariance $\Sigma_i(t)$ (decomposed into scaling $s_i(t)$ and rotation $q_i(t)$), opacity $\alpha_i(t)$, and color $c_i(t)$.

Rendering an image at time $t$ proceeds by projecting these Gaussians onto the 2D image plane. The pixel color $C(v,t)$ is computed via differentiable alpha-blending of sorted Gaussians:
\begin{equation}
C(v,t) = \sum_{i \in \mathcal{N}} c_i(t)\alpha_i(t) \prod_{j=1}^{i-1}(1 - \alpha_j(t)),
\end{equation}
where $c_i(t)$ and $\alpha_i(t)$ are the color and opacity of i-th Gaussian. This differentiable formulation allows for real-time, high-fidelity reconstruction of dynamic scenes, serving as the foundation for our open-vocabulary segmentation.

\subsection{Overview}
We address the problem of open-vocabulary language grounding in dynamic 4D scenes. Given a scene represented by a pre-trained 4D Gaussian Splatting (4DGS) model, our goal is to produce a complete spatio-temporal segmentation that corresponds to an arbitrary natural language query $\mathcal{T}$. To achieve this, we introduce PanopticQuery, a framework that performs reasoning at query time. This design allows us to interpret complex, context-dependent queries that are beyond the scope of methods relying on static, pre-computed semantic features. Our pipeline, illustrated in Fig.~\ref{fig:pipeline}, proceeds in three stages:
\begin{enumerate}
    \item Multi-View Semantic Grounder: We leverage a large foundation model's advanced reasoning to interpret the query in the context of rendered videos, generating initial 2D segmentation proposals.
    \item Multi-View Semantic Consensus: We introduce a geometric voting algorithm to transform the potentially noisy 2D proposals into geometrically reliable evidence.
    \item 4D Grounding Lifting: We use this reliable evidence to train a lightweight neural field, lifting the 2D understanding into a complete and temporally coherent 4D grounding results.
\end{enumerate}

\subsection{Multi-View Semantic Grounder}
The first stage's objective is to translate the abstract language query into concrete visual evidence. We begin by rendering RGB and depth video streams, $\{\text{Vid}_v, \text{Depth}_v\}^{V}_{v=1}$, from $V$ diverse viewpoints of the 4DGS scene.

Unlike methods that rely on matching static features (e.g., CLIP), we employ a multimodal large language model as our semantic grounder to analyze these videos. It is capable of understanding the complex semantics of actions (``the hand \textit{picking up} a cup''), relationships (``the book \textit{to the left of} the monitor''), and events that unfold over time. This is precisely the capability needed to overcome the limitations of the ``embed-first'' paradigm of previous methods. For each view $v$, the semantic grounder processes the video $\text{Vid}_v$ and query $\mathcal{T}$ to produce a sequence of 2D binary masks, $M_v = \{m_{v,t}\}$.

These multi-view masks are treated as initial candidate proposals. While the grounder excel at reasoning, they are not perfect geometric engines. The outputs can suffer from view-dependent errors, hallucinations, or inaccuracies at object boundaries, necessitating the subsequent verification stage.

\subsection{Multi-View Semantic Consensus}
This stage is critical for bridging the gap between the 2D high-level semantic understanding and the precise geometry of the 4D scene. As illustrated in Fig.~\ref{fig:pipeline}, its purpose is to filter the noisy 2D proposals and distill a high-confidence set of masks that are geometrically consistent.

Our method leverages the principle that a correct segmentation in one view must be geometrically consistent across other views, which we implement as an automated voting algorithm. An initial visibility test filters out views without sufficient correspondence in other perspectives. Specifically, for each candidate mask sequence \(M_i\) and every other sequence \(M_j\), we assess consistency at frame \(t\) by back-projecting the mask \(m_{i,t}\) (using depth \(\text{Depth}_{i,t}\)) into a 3D point cloud and re-projecting it into view \(j\) to obtain \(\Pi_{i \to j}(m_{i,t})\). Visibility is confirmed if the projected mask shares over \(60\%\) pixel overlap with \(m_{j,t}\).

We then score consistency hierarchically based on the Intersection over Union (IoU) between the projected mask $\Pi_{i \to j}(m_{i,t})$ and the proposal $m_{j,t}$:
\begin{enumerate}
    \item Frame-level Consensus: A high IoU ($> \delta$) marks consensus for a frame.
    \item View-pair Voting: If the fraction of consensus frames between views $i$ and $j$ exceeds a threshold $\epsilon$, view $j$ casts a ``vote'' for $M_i$.
    \item Global Reliability: A mask sequence $M_i$ is deemed reliable if its normalized vote count surpasses a confidence threshold $\tau$.
\end{enumerate}
This process effectively prunes proposals that are mere 2D artifacts or hallucinations, yielding a set of masks $\{M_{\text{reliable}}\}$ that represent robust, multi-view consistent evidence of the object's presence.

\subsection{4D Grounding Lifting}
\label{sec:feature_field}
The final challenge is to lift the sparse, 2D supervision from $\{M_{\text{reliable}}\}$ into a dense, continuous, and temporally consistent 4D grounding result. A simple per-frame labeling would fail to ensure temporal coherence or handle non-rigid motion smoothly. Adopting the strategy proposed in SA4D~\cite{ji2024segment}, we learn a neural field over the canonical space of the 4DGS model, keeping the original geometry and appearance frozen.

\subsubsection{Temporal Identity Feature Field.}
We leverage a lightweight MLP, $\phi_\theta$, to encode identity features. As defined in SA4D, given the center position $\mathcal{X}$ and timestamp $t$, the network predicts a time-variant identity embedding $e$ for each Gaussian:
\begin{equation}
e = \phi_{\theta}(\gamma(\mathcal{X}), \gamma(t)),
\end{equation}
where $\gamma(\cdot)$ denotes the positional encoding. To align these 3D embeddings with 2D observations, we employ differentiable feature splatting. The 2D identity feature $E_p$ at pixel $p$ is rendered via:
\begin{equation}
E_p = \sum_{i \in \mathcal{N}} e_i \alpha_i \prod_{j=1}^{i-1} (1 - \alpha_j),
\end{equation}
where $\alpha_i$ is the opacity of the $i$-th Gaussian. Subsequently, a tiny convolutional decoder $\phi_c$ and a softmax function are used to predict the Gaussian identity $f$:
\begin{equation}
f = softmax(\phi_c(E_p)).
\end{equation}

\subsubsection{Optimization.}
Following SA4D~\cite{ji2024segment} and Gaussian Grouping~\cite{ye2024gaussian},the training is supervised by a composite loss $L = \lambda_{2d}L_{2d} + \lambda_{3d}L_{3d}$. The 2D Identity Loss $L_{2d}$ is the standard cross-entropy loss on the training image $I$:
\begin{equation}
L_{2d} = -\frac{1}{\|I\|} \sum_{i \in I} \sum_{c=1}^{C} p(i) \log \hat{p}(i),
\end{equation}
where $C$ is the total number of mask identities (in our binary query setting, $C=2$), and $p$ and $\hat{p}$ denote the multi-identity probability of the ground truth mask and our network predictions, respectively.

To address sparsity and occlusions, we incorporate the 3D Regularization Loss $L_{3d}$. It is formulated as the KL divergence loss, constraining the semantic prediction $f_j$ of the $j$-th Gaussian to be similar to that of its neighbors $f_i$:
\begin{equation}
L_{3d} = \frac{1}{m} \sum_{j=1}^{m} D_{KL}(P||Q) = \frac{1}{mk} \sum_{j=1}^{m} \sum_{i=1}^{k} f_j \log \frac{f_j}{f_i},
\end{equation}
where $m$ is the number of sampled Gaussians, $P$ is the distribution of the $j$-th Gaussian, and $Q$ is the set of its $k$-nearest neighbors in the deformed space. This regularization effectively propagates supervision to occluded regions, ensuring a coherent 4D semantic field.

Once trained, the field $\phi_\theta$ implicitly encodes the complete 4D grounding. Through feature splatting, we can synthesize segmentation masks for any arbitrary viewpoint and timestamp. This process directly yields precise foreground predictions, thereby finalizing the response to the language query.

%% file: sec/5_experiment.tex
\section{Experiments}
\label{sec:experiments}

\subsection{Experimental Setup}

\subsubsection{Datasets.}
To provide a comprehensive evaluation of PanopticQuery, we perform evalution on both existing benchmark and our proposed benchmark.

(1) Neu3D~\cite{li2022neural}: We utilize the Neu3D dataset, a standard benchmark for dynamic scenes, to validate fundamental query capabilities. As scenes of Neu3D primarily feature distinct objects without complex inter-object interactions, we use it to benchmark attribute queries, e.g., ``the blue cap''. 

(2) Panoptic-L4D: As detailed in Sec.~\ref{sec:dataset}, Panoptic-L4D is explicitly constructed to contain ``resolvable ambiguities''—scenarios where objects share visual attributes but differ in behavior or location. We use this dataset to rigorously evaluate complex reasoning, such as action, interaction, and spatial queries.

\subsubsection{Evaluation Metrics.}
Following the established protocols in 4D LangSplat~\cite{li20254d}, we employ two primary metrics to comprehensively quantify the query performance: mean intersection over union (mIoU) and mean accuracy (mAcc).

Mean Intersection over Union (mIoU):
As the gold standard for semantic segmentation, mIoU measures the precise overlap between the predicted region and the ground truth. For a given natural language query, we compute the IoU frame-by-frame and report the average across all valid test frames $\mathcal{T}$:
\begin{equation}
\text{mIoU} = \frac{1}{|\mathcal{T}|} \sum_{t \in \mathcal{T}} \frac{|M_{\text{pred}}^{(t)} \cap M_{\text{gt}}^{(t)}|}{|M_{\text{pred}}^{(t)} \cup M_{\text{gt}}^{(t)}|},
\end{equation}
where $M_{\text{pred}}^{(t)}$ and $M_{\text{gt}}^{(t)}$ denote the predicted binary foreground mask and the manually annotated ground truth at frame $t$, respectively. This metric strictly penalizes false positives and false negatives, reflecting the model's capability in precise spatial localization.

Mean Accuracy (mAcc):
To assess the robustness of \textit{object-level recognition} rather than mere pixel classification, we utilize Mean Accuracy (also referred to as Frame Accuracy). This metric calculates the percentage of test frames where the model successfully localizes the target object (i.e., the predicted mask corresponds to the ground truth object). It is formulated as:
\begin{equation}
\text{mAcc} = \frac{1}{|\mathcal{Q}|} \sum_{q \in \mathcal{Q}} \frac{N_{\text{correct}}^{(q)}}{N_{\text{total}}^{(q)}},
\end{equation}
where $\mathcal{Q}$ is the set of all queries, $N_{\text{total}}^{(q)}$ is the total number of annotated frames for query $q$, and $N_{\text{correct}}^{(q)}$ represents the number of frames where the predicted mask has a valid intersection with the ground truth (indicating successful retrieval). A higher mAcc signifies the model's stability in continuously tracking and identifying the semantic target across dynamic sequences.

\subsubsection{Baselines.}
We compare PanopticQuery against two representative methods of open-vocabulary 3D/4D grounding:
(1) LangSplat~\cite{qin2024langsplat}: To evaluate static methods applied to dynamic data, we adapt the state-of-the-art 3D method, LangSplat, to operate frame-by-frame. This baseline treats the 4D scene as a sequence of independent 3D snapshots. 

(2) 4D LangSplat~\cite{li20254d}: As the current state-of-the-art for language-driven 4D grounding, this method represents the pre-embedding paradigm. It operates by embedding semantic features into Gaussian primitives during training.

\subsection{Implementation Details}
\label{sec:implementation}

All experiments were conducted on a single NVIDIA A800 GPU. 
Model Configuration. We utilize pre-trained 4DGS models as the backbone. At inference, we render $V=8$ synchronized video streams (matching the training resolution) to provide multi-view context. For the semantic grounder, we employ Qwen3-VL (8B)~\cite{Qwen3-VL} to generate initial proposals, coupled with the Samwise~\cite{cuttano2025samwise} model for segmentation.

In the geometric voting stage (Sec.~\ref{sec:method}), we empirically set the frame consistency threshold $\delta=0.3$, the temporal consensus ratio $\epsilon=0.5$, and the view reliability threshold $\tau=0.3$. 
For the final 4D Grounding Lifting, we freeze the 4DGS geometry and optimize only the feature field $\phi_\theta$ and decoder $\phi_c$. We adopt the exact hyperparameter configuration from SA4D~\cite{ji2024segment}, setting $\lambda_{2d} = 1$, $\lambda_{3d} = 2$, $m = 1000$, and $k = 5$. We use the Adam optimizer for both the temporal identity feature field network and the convolutional layer, with 2000 training iterations and a learning rate of $5 \times10^{-4}$. As the segmentation could be processed in parallel, the query time for each prompt can be reduced to around 20 minutes. Meanwhile, our final rendering speed is almost the same as that of the original 4DGS, and is essentially unaffected by the number of groups we embed. In contrast, 4DLangSplat requires about 20 hours of preprocessing for each scene, whereas our method requires no preprocessing.

\subsection{Quantitative Analysis}

\subsubsection{Evaluation on Neu3D}
We extended our evaluation on Neu3D to cover all four query dimensions, attribute, action, spatial, and interaction, to test generalization. As detailed in Tab.~\ref{tab:Neu3D_comparison}, our proposed PanopticQuery demonstrates superior performance against the other competitors over all metrics, confirming that our query-time paradigm enable the model with a universal understanding about the query and scene.
However, it is important to note that the scene in the Neu3D dataset inherently lacks ``resolvable ambiguity'' (i.e., most objects are distinguishable by appearance alone). Consequently, we further evaluate all the methods with the Panoptic-L4D benchmark.

\input{table/Neu3D_comparison}
\input{table/Panotic_L4D_comparison}

\input{picture/vis3/coffee/coffee_jpg}
\input{picture/vis3/birds/birds_jpg}

\subsubsection{Evaluation on Panoptic-L4D}
The core advantage of our framework is demonstrated on the Panoptic-L4D benchmark, which consists of various complex queries and requires comprehensive awareness of the whole scene. As shown in Tab.~\ref{tab:panotic_L4D_comparison}, PanopticQuery significantly outperforms the other two method across all complex query categories.

1) Superiority in Complex Queries:
The most striking observation is the substantial performance gap in action, spatial, and interaction queries. As hypothesized, the pre-embedding paradigm (4D LangSplat) struggles severely in these categories, achieving mAcc scores of only 34.72\%, 21.26\%, and 28.56\%, respectively. This empirical evidence confirms that static semantic features (e.g., CLIP embedding) baked into Gaussians are fundamentally insufficient for capturing dynamic behaviors or relative spatial configurations. 
In contrast, PanopticQuery leverages the dynamic reasoning capabilities of semantic grounder at query time, achieving remarkable improvements. Specifically, we observe performance jumps of +48.9\% in action queries, +59.9\% in spatial queries, and +41.3\% in interaction queries (mAcc). This validates that our framework successfully resolves ambiguities that rely on motion cues (e.g., distinguishing ``sitting'' from ``standing'') and relational context (e.g., distinguishing ``held by hand'' from ``keep still''), which are remain challenging for prior art for prior arts.

2) Robustness in Attribute Queries:
For standard attribute queries (e.g., color, object category), our method maintains a solid lead (76.72\% mAcc vs. 71.78\% for 4D LangSplat). While the gap is narrower compared to complex queries, this result is crucial: it demonstrates that adopting a dynamic, query-time architecture does not compromise the model's ability to handle fundamental static recognition tasks.

Overall, these results strongly substantiate our core contribution: shifting from ``static pre-embedding'' to ``dynamic query-time reasoning'' is the key to unlocking true open-vocabulary understanding in 4D dynamic scenes.

\subsection{Qualitative Comparison}
We first present qualitative results on the Neu3D dataset, as shown in Fig.~\ref{fig:qualitative_neu}. From the figure, it is evident that both LangSplat and 4D LangSplat struggle to accurately interpret the scene. For example, they often localize multiple unrelated objects when prompted with a query involving “cup”, resulting in incorrect or ambiguous grounding. Besides, they fails to capture the true subject of query, such as "The cap". In contrast, our proposed PanopticQuery demonstrates a stronger contextual understanding, enabling precise identification and localization of the target object. It successfully finds the right object (\textit{e.g.}, "the pot") by analyzing the contextual relationship between "pot" and "man" in the scene.

We further showcase results on Panoptic-L4D, a more challenging benchmark featuring complex and diverse query types. As illustrated in Fig.~\ref{fig:qualitative_pano}, LangSplat produces temporally inconsistent outputs due to its lack of temporal modeling, and 4D LangSplat fails to ground the correct objects in multi-entity queries. In comparison, PanopticQuery successfully identifies the intended targets by leveraging spatio-temporal context and reasoning at query time. Specifically, for spatial query such as "the bucket" in Q1 of the second scene, previous methods can not distinguish the relative position of the three buckets, leading to wrong results. Instead, PanopticQuery is capable of reasoning about the spatial relationships among them and ground the target bucket. Furthermore, we demonstrate the ability of PanopticQuery for diverse query types in Fig.~\ref{fig:qualitative_mouse_cup}, including attribute query, action query, spatial query, and interaction query.

These results highlight the effectiveness of our framework in achieving comprehensive, panoptic scene understanding. Through dynamic reasoning, our approach handles universal queries with complex semantics more robustly than prior methods. \textit{Additional qualitative examples can be found in the supplementary material and demo video.}

\input{picture/vis2/mouse_cup/mouse_cup_vis}

\revised{\subsection{Composite Query}
\input{table/quantitative_composition}
Beyond single-type queries, our method also supports composite queries that combine various query types. We evaluate five composite queries on both the Neu3D and Panoptic-L4D datasets, as shown in Tab.~\ref{tab:combination_query_results}. These results indicate that our method is capable of understanding complex queries and locating target objects.}

\subsection{Ablation Studies}

We conduct extensive ablation studies on the Panoptic-L4D dataset to validate the contribution of each component in our pipeline.

\subsubsection{Ablation on Multi-View Semantic Consensus}
\input{table/ablation_no_vote}
To evaluate the contribution of the multi-view semantic consensus module, we compared the results obtained by lifting all candidate 2D masks into 4D space. As indicated in Tab.~\ref{tab:ablation_no_vote}, the absence of the multi-view semantic consensus module leads to performance degradation, primarily due to the inherent spatiotemporal inconsistencies among the independently estimated 2D masks across video frames. Conversely, by integrating this module, we effectively resolve potential inter-mask conflicts and achieve a statistically significant improvement in performance.

\subsubsection{Ablation on Number of View.}
\input{table/view_nums_ablation}
We investigate the trade-off between geometric coverage and computational efficiency by varying the number of rendered views $V \in \{\revised{1}, 4, 6, 8, 10, 12\}$. As detailed in Tab.~\ref{tab:ablation_cams}, the impact of view density varies significantly across query types. 

For attribute queries, performance remains relatively robust even at lower densities (e.g., $\sim$78\% mAcc at $V=4$), suggesting that intrinsic object properties are often discernible from sparse viewpoints. However, for complex reasoning tasks involving action, spatial, and interaction, sparse configurations ($V \le 6$) fail to provide sufficient visual overlap for our geometric voting mechanism, leading to unresolved occlusions and ambiguities. 
\revised{For $V=1$, we find that complex queries suffered massive drops, which clearly demonstrates the necessity of multi-view input and MVSC.}
Crucially, increasing $V$ from 4 to 8 yields a substantial performance leap; specifically, action accuracy increases from 67.48\% to 83.66\%, and spatial accuracy improves by nearly 20\%. Beyond this point ($V > 8$), performance gains saturate or even slightly degrade due to diminishing returns in visual information and potential noise accumulation. Consequently, we adopt $V=8$ as the optimal configuration that balances high-fidelity segmentation with reasonable inference cost.

\subsubsection{Ablation on Vision-Language Models.} 
\input{table/ablation_vlm}

To verify the robustness and model-agnostic nature of our approach, we evaluate performance using various state-of-the-art VLMs, as shown in Tab.~\ref{tab:ablation_vlm}. 

First, the w/o VLM baseline (which relies solely on the segmentation model's internal text encoder without VLM reasoning) shows a noticeable performance drop across all metrics, validating the necessity of an explicit reasoning stage for disambiguation. 
Second, when equipping our framework with different advanced VLMs, ranging from Qwen2.5-VL to closed-source giants like GPT-5.2 and Gemini3-Pro, we observe consistently high performance. This indicates that our PanopticQuery pipeline is robust to different VLMs and effectively leverages the potential of VLMs.

\subsubsection{Performance in Scenarios of Different Scales}

\input{table/quantitative_all_scene_scale}
\revised{We evaluate the performance of our method across scenes of various scales. As shown in Tab.~\ref{tab:avg_scene_results}, our method exhibits robust performance across different spatial scales, indicating strong generalization capabilities in diverse environments.}

%% file: table/Neu3D_comparison.tex
\begin{table}[t]
\centering
\caption{Quantitative comparison of different query types on Neu3D. Our method achieves superior performance across all categories. Cells highlighted in \textbf{red} and \textbf{yellow} denote the best and second-best performance within each query category.}
\vspace{-3mm}
\label{tab:Neu3D_comparison}

\setlength{\tabcolsep}{1.0pt} 
{
    
    \begin{tabular}{lcccccccc}
    \toprule\hline
    \multirow{2.5}{*}{Method} & \multicolumn{2}{c}{\textbf{Attr. Query}} & \multicolumn{2}{c}{\textbf{Act. Query}} & \multicolumn{2}{c}{\textbf{Spat. Query}} & \multicolumn{2}{c}{\textbf{Inter. Query}} \\
    \cmidrule(lr){2-3} \cmidrule(lr){4-5} \cmidrule(lr){6-7} \cmidrule(lr){8-9}
     & mAcc & mIoU & mAcc & mIoU & mAcc & mIoU & mAcc & mIoU \\
    \midrule

    \rowcolor{gray!20}
    LangSplat    & 69.45 & 42.14 & 53.33 & 42.42 & 26.67 & 17.16 & 21.42 & 12.13 \\

    4D LangSplat & \cellcolor{yellow!22}84.60 & \cellcolor{yellow!22}68.73 & \cellcolor{yellow!22}83.33 & \cellcolor{yellow!22}70.15 & \cellcolor{yellow!22}53.33 & \cellcolor{yellow!22}43.57 & \cellcolor{yellow!22}57.14 & \cellcolor{yellow!22}42.93 \\

    \midrule 
    \rowcolor{gray!20}
    Ours & \cellcolor{red!18}89.55 & \cellcolor{red!18}75.47 & \cellcolor{red!18}91.67 & \cellcolor{red!18}76.52 & \cellcolor{red!18}73.33 & \cellcolor{red!18}62.27 & \cellcolor{red!18}78.57 & \cellcolor{red!18}67.41 \\
    \hline\bottomrule
    \end{tabular}
}
\vspace{-3mm}
\end{table}

%% file: table/Panotic_L4D_comparison.tex
\begin{table}[t]
\centering
\caption{Quantitative comparison of query types on Panoptic-L4D dataset. Our method significantly outperforms other methods in dynamic and complex scenarios. Cells highlighted in \textbf{red} and \textbf{yellow} denote the best and second-best performance within each query category.}
\vspace{-3mm}
\label{tab:panotic_L4D_comparison}

\setlength{\tabcolsep}{1.0pt} 
{
    
    \begin{tabular}{lcccccccc}
    \toprule\hline
    \multirow{2.5}{*}{Method} & \multicolumn{2}{c}{\textbf{Attr. Query}} & \multicolumn{2}{c}{\textbf{Act. Query}} & \multicolumn{2}{c}{\textbf{Spat. Query}} & \multicolumn{2}{c}{\textbf{Inter. Query}} \\
    \cmidrule(lr){2-3} \cmidrule(lr){4-5} \cmidrule(lr){6-7} \cmidrule(lr){8-9}
     & mAcc & mIoU & mAcc & mIoU & mAcc & mIoU & mAcc & mIoU \\
    \midrule

    \rowcolor{gray!20}
    LangSplat    & 47.79 & 28.81 & 18.27 & 12.72 & 10.12 & 6.61  & 16.03 & 11.47 \\

    4D LangSplat & \cellcolor{yellow!22}71.78 & \cellcolor{yellow!22}54.30 & \cellcolor{yellow!22}34.72 & \cellcolor{yellow!22}29.00 & \cellcolor{yellow!22}21.26 & \cellcolor{yellow!22}15.13 & \cellcolor{yellow!22}28.56 & \cellcolor{yellow!22}22.84 \\

    \midrule 
    \rowcolor{gray!20}
    Ours & \cellcolor{red!18}76.72 & \cellcolor{red!18}67.11 & \cellcolor{red!18}83.66 & \cellcolor{red!18}75.14 & \cellcolor{red!18}81.19 & \cellcolor{red!18}72.97 & \cellcolor{red!18}69.83 & \cellcolor{red!18}63.73 \\
    \hline\bottomrule
    \end{tabular}
}
\vspace{-3mm}
\end{table}

%% file: picture/vis3/coffee/coffee_jpg.tex
\definecolor{bgA}{RGB}{223, 241, 215} 
\definecolor{bgB}{RGB}{219, 236, 249} 
\definecolor{bgC}{RGB}{255, 235, 235} 

\newcommand{\basepathG}{picture/vis3/coffee/the_cap_on_the_man_head_jpg/}
\newcommand{\basepathH}{picture/vis3/coffee/the_cup_held_by_the_man_jpg/}

\newcommand{\basepathI}{picture/vis3/cook/the_man_cooking_jpg/}
\newcommand{\basepathJ}{picture/vis3/cook/the_pot_being_used_by_the_man_to_cook_spinach_jpg/}

\newtcolorbox{groupbox}[2][]{
    enhanced, colback=#2, colframe=#2, arc=2mm, boxrule=0pt,
    left=1pt, right=1pt, top=2pt, bottom=1pt, boxsep=0pt,
    width=\linewidth, #1
}

\newlength{\myImgW}
\setlength{\myImgW}{0.1044\textwidth} 

\newcommand{\imgCell}[1]{\includegraphics[width=\myImgW]{#1}}
\newcommand{\imgRGB}[1]{\includegraphics[width=\myImgW]{#1}}

\newcommand{\titleBox}[1]{%
    \parbox[c][0.85cm][c]{\linewidth}{%
        \centering \small \bfseries \linespread{0.9}\selectfont #1%
    }%
}

\begin{figure*}[t]
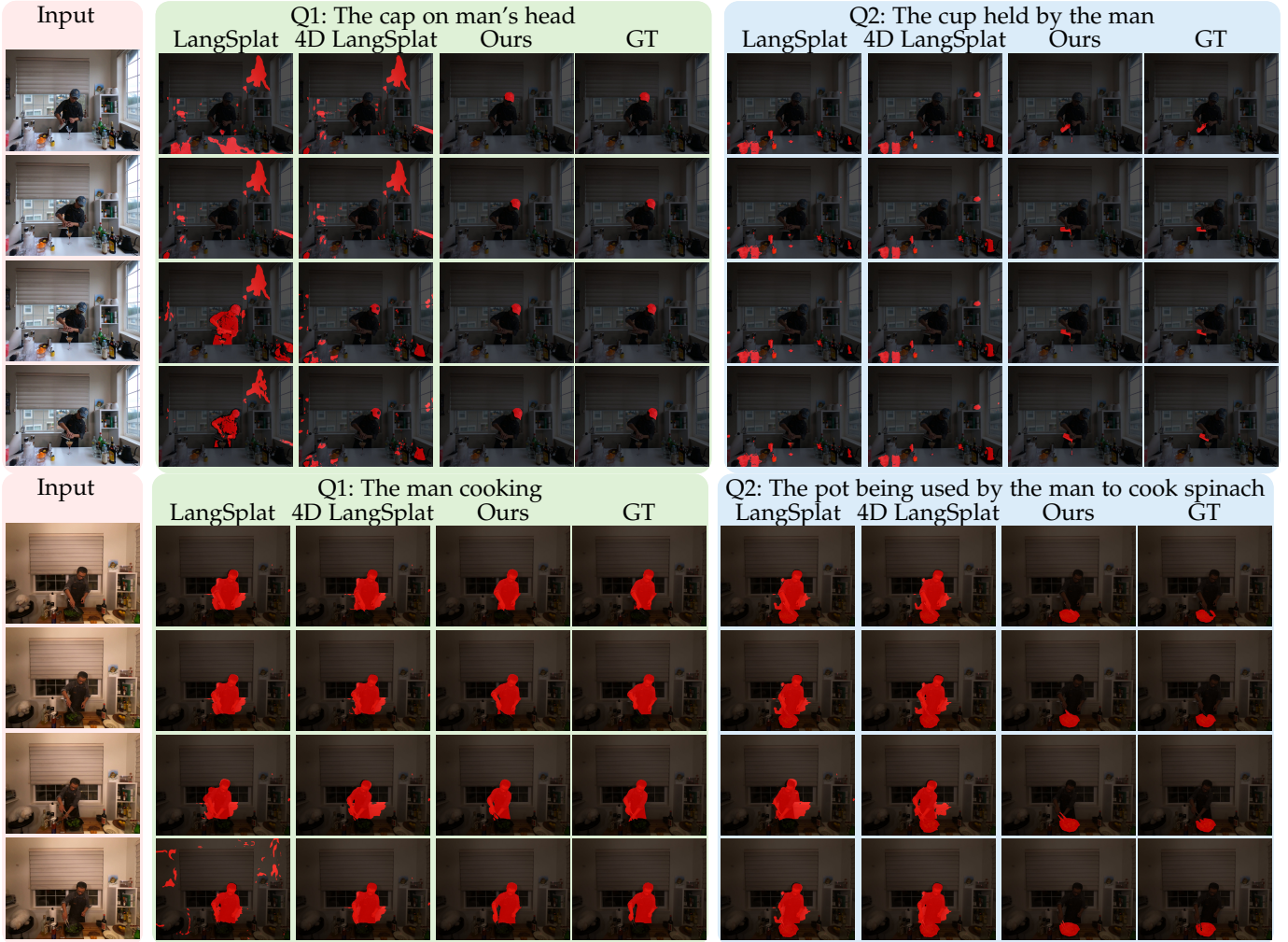

    \centering

    \begin{minipage}[t]{0.11\linewidth}
        \begin{groupbox}[width=\linewidth, colback=bgC, arc=2mm]{bgC}
            \centering
            Input
            \setlength{\tabcolsep}{0pt} 
            \renewcommand{\arraystretch}{0.6}
            
            \begin{tabular}{c}
                \phantom{ 3D} \\ 
                \imgRGB{\basepathG 0000.jpg} \\
                \imgRGB{\basepathG 0080.jpg} \\
                \imgRGB{\basepathG 0160.jpg} \\
                \imgRGB{\basepathG 0240.jpg} 
            \end{tabular}
        \end{groupbox}
    \end{minipage}
    \hfill 
    \begin{minipage}[t]{0.435\linewidth}
        \begin{groupbox}{bgA}
            \centering
            Q1: The cap on man's head
            
            \setlength{\tabcolsep}{0.5pt} 
            \renewcommand{\arraystretch}{0.5} 
            
            \begin{tabular}{cccc}
                 LangSplat &  4D LangSplat &  Ours &  GT \\
                \imgCell{\basepathG d3_0.jpg} & \imgCell{\basepathG d4_0.jpg} & \imgCell{\basepathG pred_0.jpg} & \imgCell{\basepathG gt_0.jpg} \\
                \imgCell{\basepathG d3_80.jpg} & \imgCell{\basepathG d4_80.jpg} & \imgCell{\basepathG pred_80.jpg} & \imgCell{\basepathG gt_80.jpg} \\
                \imgCell{\basepathG d3_160.jpg} & \imgCell{\basepathG d4_160.jpg} & \imgCell{\basepathG pred_160.jpg} & \imgCell{\basepathG gt_160.jpg} \\
                \imgCell{\basepathG d3_240.jpg} & \imgCell{\basepathG d4_240.jpg} & \imgCell{\basepathG pred_240.jpg} & \imgCell{\basepathG gt_240.jpg} \\
            \end{tabular}
        \end{groupbox}
    \end{minipage}
    \hfill 
    \begin{minipage}[t]{0.435\linewidth}
        \begin{groupbox}{bgB}
            \centering
            Q2: The cup held by the man
            \setlength{\tabcolsep}{0.5pt} 
            \renewcommand{\arraystretch}{0.5}  
            \begin{tabular}{cccc}
                 LangSplat &  4D LangSplat &  Ours &  GT \\
                \imgCell{\basepathH d3_0.jpg} & \imgCell{\basepathH d4_0.jpg} & \imgCell{\basepathH pred_0.jpg} & \imgCell{\basepathH gt_0.jpg} \\
                \imgCell{\basepathH d3_80.jpg} & \imgCell{\basepathH d4_80.jpg} & \imgCell{\basepathH pred_80.jpg} & \imgCell{\basepathH gt_80.jpg} \\
                \imgCell{\basepathH d3_160.jpg} & \imgCell{\basepathH d4_160.jpg} & \imgCell{\basepathH pred_160.jpg} & \imgCell{\basepathH gt_160.jpg} \\
                \imgCell{\basepathH d3_240.jpg} & \imgCell{\basepathH d4_240.jpg} & \imgCell{\basepathH pred_240.jpg} & \imgCell{\basepathH gt_240.jpg} \\
            \end{tabular}
        \end{groupbox}
    \end{minipage}
    \\

    \begin{minipage}[t]{0.11\linewidth}
        \begin{groupbox}[width=\linewidth, colback=bgC, arc=2mm]{bgC}
            \centering
            Input
            \setlength{\tabcolsep}{0pt} 
            \renewcommand{\arraystretch}{0.6}
            
            \begin{tabular}{c}
                \phantom{ 3D} \\ 
                \imgRGB{\basepathI 0000.jpg} \\
                \imgRGB{\basepathI 0080.jpg} \\
                \imgRGB{\basepathI 0160.jpg} \\
                \imgRGB{\basepathI 0240.jpg} 
            \end{tabular}
        \end{groupbox}
    \end{minipage}
    \hfill 
    \begin{minipage}[t]{0.435\linewidth}
        \begin{groupbox}{bgA}
            \centering
            Q1: The man cooking
            
            \setlength{\tabcolsep}{0.5pt} 
            \renewcommand{\arraystretch}{0.5} 
            
            \begin{tabular}{cccc}
                 LangSplat &  4D LangSplat &  Ours &  GT \\
                \imgCell{\basepathI d3_0.jpg} & \imgCell{\basepathI d4_0.jpg} & \imgCell{\basepathI pred_0.jpg} & \imgCell{\basepathI gt_0.jpg} \\
                \imgCell{\basepathI d3_80.jpg} & \imgCell{\basepathI d4_80.jpg} & \imgCell{\basepathI pred_80.jpg} & \imgCell{\basepathI gt_80.jpg} \\
                \imgCell{\basepathI d3_160.jpg} & \imgCell{\basepathI d4_160.jpg} & \imgCell{\basepathI pred_160.jpg} & \imgCell{\basepathI gt_160.jpg} \\
                \imgCell{\basepathI d3_240.jpg} & \imgCell{\basepathI d4_240.jpg} & \imgCell{\basepathI pred_240.jpg} & \imgCell{\basepathI gt_240.jpg} \\
            \end{tabular}
        \end{groupbox}
    \end{minipage}
    \hfill 
    \begin{minipage}[t]{0.435\linewidth}
        \begin{groupbox}{bgB}
            \centering
            Q2: The pot being used by the man to cook spinach
            
            \setlength{\tabcolsep}{0.5pt} 
            \renewcommand{\arraystretch}{0.5} 
            
            \begin{tabular}{cccc}
                 LangSplat &  4D LangSplat &  Ours &  GT \\
                \imgCell{\basepathJ d3_0.jpg} & \imgCell{\basepathJ d4_0.jpg} & \imgCell{\basepathJ pred_0.jpg} & \imgCell{\basepathJ gt_0.jpg} \\
                \imgCell{\basepathJ d3_80.jpg} & \imgCell{\basepathJ d4_80.jpg} & \imgCell{\basepathJ pred_80.jpg} & \imgCell{\basepathJ gt_80.jpg} \\
                \imgCell{\basepathJ d3_160.jpg} & \imgCell{\basepathJ d4_160.jpg} & \imgCell{\basepathJ pred_160.jpg} & \imgCell{\basepathJ gt_160.jpg} \\
                \imgCell{\basepathJ d3_240.jpg} & \imgCell{\basepathJ d4_240.jpg} & \imgCell{\basepathJ pred_240.jpg} & \imgCell{\basepathJ gt_240.jpg} \\
            \end{tabular}
        \end{groupbox}
    \end{minipage}
    \vspace{-3mm}
    \caption{Qualitative Comparison on Neu3D dataset.}
    \vspace{-3mm}
    \label{fig:qualitative_neu}
\end{figure*}

%% file: picture/vis3/birds/birds_jpg.tex
\definecolor{bgA}{RGB}{223, 241, 215} 
\definecolor{bgB}{RGB}{219, 236, 249} 
\definecolor{bgC}{RGB}{255, 235, 235} 

\newcommand{\basepathA}{picture/vis3/birds/the_bird_flying_down_from_the_wooden_dock_jpg/}
\newcommand{\basepathB}{picture/vis3/birds/the_bird_standing_by_the_water_dish_jpg/}
\newcommand{\basepathC}{picture/vis3/buckets/the_bucket_between_the_red_and_blue_buckets_jpg/}
\newcommand{\basepathD}{picture/vis3/buckets/the_bucket_which_is_lifted_by_the_woman_in_pink_jpg/}


\setlength{\myImgW}{0.1044\textwidth} 



\begin{figure*}[t]
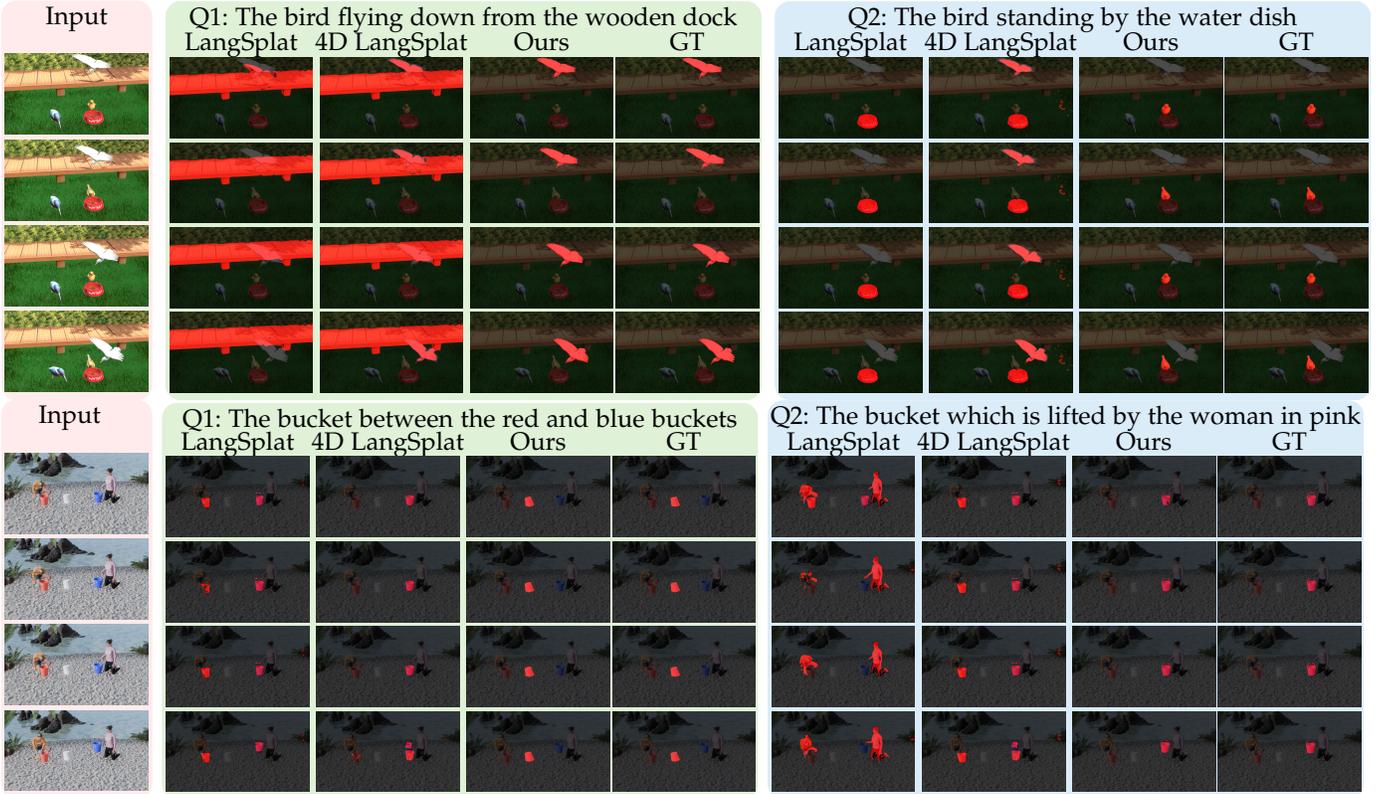

    \centering
    
    \begin{minipage}[t]{0.11\linewidth}
        \begin{groupbox}{bgC}
            \centering
            Input \\
            
            \setlength{\tabcolsep}{0pt} 
            \renewcommand{\arraystretch}{0.6}
            
            \begin{tabular}{c}
                \phantom{3D} \\ 
                \imgRGB{\basepathA frame_0001.jpg} \\
                \imgRGB{\basepathA frame_0033.jpg} \\
                \imgRGB{\basepathA frame_0065.jpg} \\
                \imgRGB{\basepathA frame_0097.jpg} 
            \end{tabular}
        \end{groupbox}
    \end{minipage}
    \hfill 
    \begin{minipage}[t]{0.435\linewidth}
        \begin{groupbox}{bgA}
            \centering
            Q1: The bird flying down from the wooden dock
            
            \setlength{\tabcolsep}{0.5pt} 
            \renewcommand{\arraystretch}{0.5} 
            
            \begin{tabular}{cccc}
                LangSplat & 4D LangSplat & Ours & GT \\
                \imgCell{\basepathA d3_0.jpg} & \imgCell{\basepathA d4_0.jpg} & \imgCell{\basepathA pred_0.jpg} & \imgCell{\basepathA gt_0.jpg} \\
                \imgCell{\basepathA d3_32.jpg} & \imgCell{\basepathA d4_32.jpg} & \imgCell{\basepathA pred_32.jpg} & \imgCell{\basepathA gt_32.jpg} \\
                \imgCell{\basepathA d3_64.jpg} & \imgCell{\basepathA d4_64.jpg} & \imgCell{\basepathA pred_64.jpg} & \imgCell{\basepathA gt_64.jpg} \\
                \imgCell{\basepathA d3_96.jpg} & \imgCell{\basepathA d4_96.jpg} & \imgCell{\basepathA pred_96.jpg} & \imgCell{\basepathA gt_96.jpg} \\
            \end{tabular}
        \end{groupbox}
    \end{minipage}
    \hfill 
    \begin{minipage}[t]{0.435\linewidth}
        \begin{groupbox}{bgB}
            \centering
            Q2: The bird standing by the water dish
            
            \setlength{\tabcolsep}{0.5pt} 
            \renewcommand{\arraystretch}{0.5} 
            
            \begin{tabular}{cccc}
                LangSplat & 4D LangSplat & Ours & GT \\
                \imgCell{\basepathB d3_0.jpg} & \imgCell{\basepathB d4_0.jpg} & \imgCell{\basepathB pred_0.jpg} & \imgCell{\basepathB gt_0.jpg} \\
                \imgCell{\basepathB d3_32.jpg} & \imgCell{\basepathB d4_32.jpg} & \imgCell{\basepathB pred_32.jpg} & \imgCell{\basepathB gt_32.jpg} \\
                \imgCell{\basepathB d3_64.jpg} & \imgCell{\basepathB d4_64.jpg} & \imgCell{\basepathB pred_64.jpg} & \imgCell{\basepathB gt_64.jpg} \\
                \imgCell{\basepathB d3_96.jpg} & \imgCell{\basepathB d4_96.jpg} & \imgCell{\basepathB pred_96.jpg} & \imgCell{\basepathB gt_96.jpg} \\
            \end{tabular}
        \end{groupbox}
    \end{minipage}
    \\
    
    \begin{minipage}[t]{0.11\linewidth}
        \begin{groupbox}[width=\linewidth, colback=bgC, arc=2mm]{bgC}
            \centering
            Input
            \setlength{\tabcolsep}{0pt} 
            \renewcommand{\arraystretch}{0.6}
            
            \begin{tabular}{c}
                \phantom{ 3D} \\ 
                \imgRGB{\basepathC frame_0001.jpg} \\
                \imgRGB{\basepathC frame_0033.jpg} \\
                \imgRGB{\basepathC frame_0065.jpg} \\
                \imgRGB{\basepathC frame_0097.jpg} 
            \end{tabular}
        \end{groupbox}
    \end{minipage}
    \hfill 
    \begin{minipage}[t]{0.435\linewidth}
        \begin{groupbox}{bgA}
            \centering
            Q1: The bucket between the red and blue buckets

            \setlength{\tabcolsep}{0.5pt} 
            \renewcommand{\arraystretch}{0.5} 
            
            \begin{tabular}{cccc}
                 LangSplat &  4D LangSplat &  Ours &  GT \\
                \imgCell{\basepathC d3_0.jpg} & \imgCell{\basepathC d4_0.jpg} & \imgCell{\basepathC pred_0.jpg} & \imgCell{\basepathC gt_0.jpg} \\
                \imgCell{\basepathC d3_32.jpg} & \imgCell{\basepathC d4_32.jpg} & \imgCell{\basepathC pred_32.jpg} & \imgCell{\basepathC gt_32.jpg} \\
                \imgCell{\basepathC d3_64.jpg} & \imgCell{\basepathC d4_64.jpg} & \imgCell{\basepathC pred_64.jpg} & \imgCell{\basepathC gt_64.jpg} \\
                \imgCell{\basepathC d3_96.jpg} & \imgCell{\basepathC d4_96.jpg} & \imgCell{\basepathC pred_96.jpg} & \imgCell{\basepathC gt_96.jpg} \\
            \end{tabular}
        \end{groupbox}
    \end{minipage}
    \hfill 
    \begin{minipage}[t]{0.435\linewidth}
        \begin{groupbox}{bgB}
            \centering
            Q2: The bucket which is lifted by the woman in pink
            
            \setlength{\tabcolsep}{0.5pt} 
            \renewcommand{\arraystretch}{0.5} 
            
            \begin{tabular}{cccc}
                 LangSplat &  4D LangSplat &  Ours &  GT \\
                \imgCell{\basepathD d3_0.jpg} & \imgCell{\basepathD d4_0.jpg} & \imgCell{\basepathD pred_0.jpg} & \imgCell{\basepathD gt_0.jpg} \\
                \imgCell{\basepathD d3_32.jpg} & \imgCell{\basepathD d4_32.jpg} & \imgCell{\basepathD pred_32.jpg} & \imgCell{\basepathD gt_32.jpg} \\
                \imgCell{\basepathD d3_64.jpg} & \imgCell{\basepathD d4_64.jpg} & \imgCell{\basepathD pred_64.jpg} & \imgCell{\basepathD gt_64.jpg} \\
                \imgCell{\basepathD d3_96.jpg} & \imgCell{\basepathD d4_96.jpg} & \imgCell{\basepathD pred_96.jpg} & \imgCell{\basepathD gt_96.jpg} \\
            \end{tabular}
        \end{groupbox}
    \end{minipage}
    \vspace{-3mm}
    \caption{Qualitative Comparison on Panoptic-L4D dataset.}
    \vspace{-3mm}
    \label{fig:qualitative_pano}
\end{figure*}

%% file: picture/vis2/mouse_cup/mouse_cup_vis.tex
\begin{figure*}[t]

    \centering
    \subfloat[Input\label{fig:qualitative_mouse_cup_input}]{%
        \includegraphics[width=0.98\textwidth]{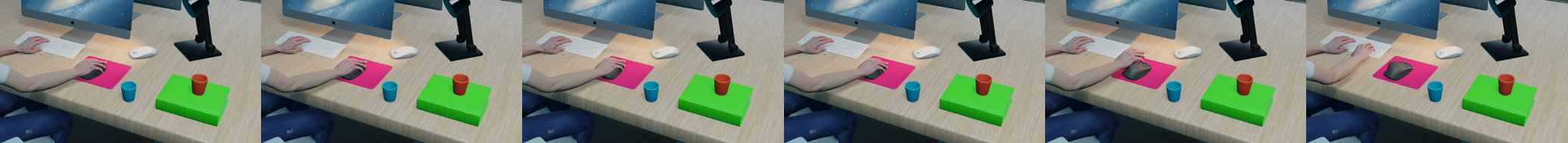}
    }\\[-0.4mm]

    \subfloat[Q1: The blue cup\label{fig:qualitative_mouse_cup_q1}]{%
        \includegraphics[width=0.98\textwidth]{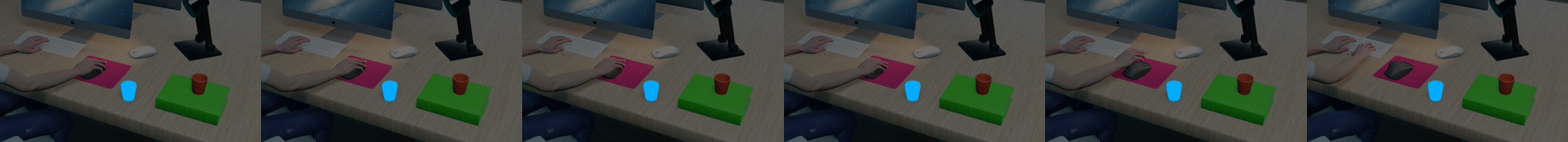}
    }\\[-0.4mm]

    \subfloat[Q2: The mouse that is lying still\label{fig:qualitative_mouse_cup_q2}]{%
        \includegraphics[width=0.98\textwidth]{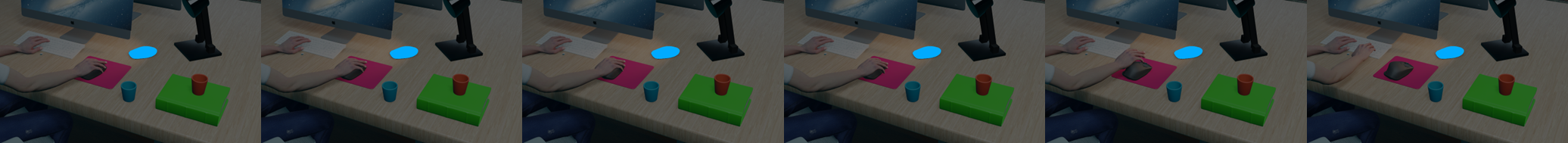}
    }\\[-0.4mm]

    \subfloat[Q3: The cup on the green book\label{fig:qualitative_mouse_cup_q3}]{%
        \includegraphics[width=0.98\textwidth]{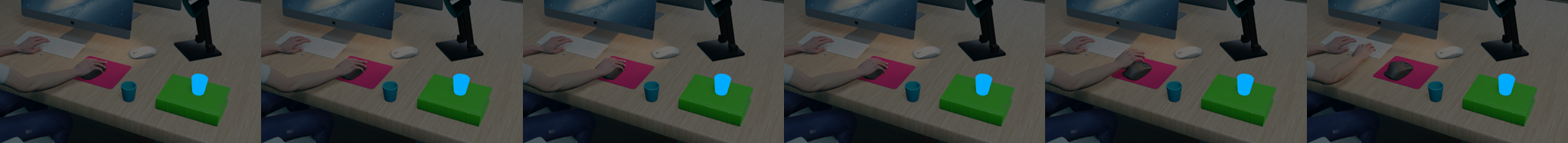}
    }\\[-0.4mm]

    \subfloat[Q4: The mouse which is moved by hand\label{fig:qualitative_mouse_cup_q4}]{%
        \includegraphics[width=0.98\textwidth]{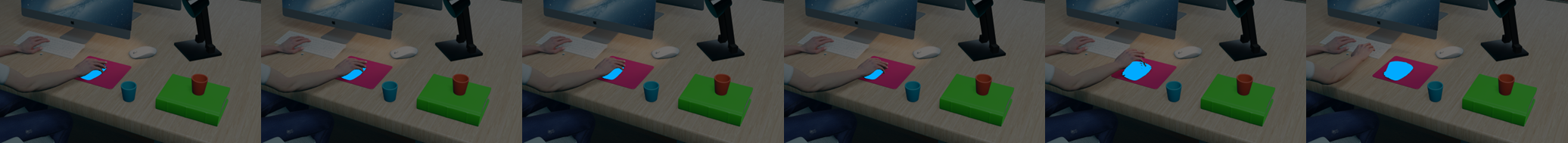}
    }\\[-0.4mm]

    \caption{Various query types' results of our method on Panoptic-L4D dataset.}
    \label{fig:qualitative_mouse_cup}
\end{figure*}

%% file: table/quantitative_composition.tex
\begin{table}[h]
\centering
\caption{Quantitative results on multi-type combination queries. We evaluate methods on five complex composite queries across two datasets.}
\vspace{-3mm}
\label{tab:combination_query_results}

\setlength{\tabcolsep}{0.8pt} 

\resizebox{\columnwidth}{!}{
    
    \begin{tabular}{lcccccccccc}
    \toprule\hline
    \multirow{2.5}{*}{Dataset} & \multicolumn{2}{c}{\textbf{Attr.\&Act.}} & \multicolumn{2}{c}{\textbf{Attr.\&Int.}} & \multicolumn{2}{c}{\textbf{Attr.\&Spat.}} & \multicolumn{2}{c}{\textbf{Spat.\&Act.}} & \multicolumn{2}{c}{\textbf{Spat.\&Int.}} \\
    \cmidrule(lr){2-3} \cmidrule(lr){4-5} \cmidrule(lr){6-7} \cmidrule(lr){8-9} \cmidrule(lr){10-11}
     & mAcc & mIoU & mAcc & mIoU & mAcc & mIoU & mAcc & mIoU & mAcc & mIoU \\
    \midrule

    Neu3D        & 91.67 & 76.47 & 83.33 & 71.26 & 88.89 & 74.06 & 100.00 & 80.61 & 66.67 & 57.93 \\
    
    \rowcolor{gray!20}
    Panoptic-L4D & 82.32 & 72.91 & 79.17 & 70.98 & 80.25 & 70.05 & 83.58  & 69.18 & 77.21 & 67.95 \\
    \hline\bottomrule
    \end{tabular}
}
\vspace{-3mm}
\end{table}

%% file: table/ablation_no_vote.tex
\begin{table}[t]
\centering
\caption{Ablation study on Multi-View Semantic Consensus (MVSC). It indicates that the proposed consensus mechanism plays a key role in producing 4D-consistent results. Cells highlighted in \textbf{red} denote the best performance within each query category.}
\vspace{-3mm}
\label{tab:ablation_no_vote}

\setlength{\tabcolsep}{1.2pt} 
{
    
    \begin{tabular}{lcccccccc}
    \toprule\hline
    \multirow{2.5}{*}{} & \multicolumn{2}{c}{\textbf{Attr. Query}} & \multicolumn{2}{c}{\textbf{Act. Query}} & \multicolumn{2}{c}{\textbf{Spat. Query}} & \multicolumn{2}{c}{\textbf{Inter. Query}} \\
    \cmidrule(lr){2-3} \cmidrule(lr){4-5} \cmidrule(lr){6-7} \cmidrule(lr){8-9}
     & mAcc & mIoU & mAcc & mIoU & mAcc & mIoU & mAcc & mIoU \\
    \midrule

    w/o MVSC      & 76.24            & \cellcolor{red!18}67.20            & 79.15           & 67.04          & 71.76                & 64.29                & 65.15             & 60.10             \\

    \rowcolor{gray!20}
    Ours & \cellcolor{red!18}76.72 & 67.11 & \cellcolor{red!18}83.66 & \cellcolor{red!18}75.14 & \cellcolor{red!18}81.19 & \cellcolor{red!18}72.97 & \cellcolor{red!18}69.83 & \cellcolor{red!18}63.73 \\
    \hline\bottomrule
    \end{tabular}
}
\vspace{-3mm}
\end{table}

%% file: table/view_nums_ablation.tex
\begin{table}[t]
\centering
\caption{Ablation study on the number of cameras. We evaluate the impact of camera density on different query types using mAcc(\%) and mIoU(\%). Cells highlighted in \textbf{red} and \textbf{yellow} denote the best and second-best performance within each query category.}
\vspace{-3mm}
\label{tab:ablation_cams}

\setlength{\tabcolsep}{2.5pt}
{
\begin{tabular}{lcccccccc}
\toprule\hline
\multirow{2.5}{*}{\# Cams} &
\multicolumn{2}{c}{\textbf{Attr. Query}} &
\multicolumn{2}{c}{\textbf{Act. Query}} &
\multicolumn{2}{c}{\textbf{Spat. Query}} &
\multicolumn{2}{c}{\textbf{Inter. Query}} \\
\cmidrule(lr){2-3} \cmidrule(lr){4-5} \cmidrule(lr){6-7} \cmidrule(lr){8-9}
& mAcc & mIoU & mAcc & mIoU & mAcc & mIoU & mAcc & mIoU \\
\midrule

1 cam
& 71.33 & 62.31
& 61.48 & 49.82
& 52.66 & 43.51
& 58.14 & 45.29 \\

\rowcolor{gray!20}
4 cams
& \cellcolor{red!18}78.62 & 66.13
& 67.48 & 53.30
& 62.66 & 47.44
& 61.18 & 46.89 \\

6 cams
& 77.18 & 68.19
& 65.41 & 52.29
& 61.04 & 50.28
& 62.22 & 48.68 \\

\rowcolor{gray!20}
8 cams
& 76.72 & 67.11
& \cellcolor{red!18}83.66 & \cellcolor{red!18}75.14
& \cellcolor{yellow!22}81.19 & \cellcolor{yellow!22}72.97
& \cellcolor{red!18}69.83 & \cellcolor{red!18}63.73 \\

10 cams
& 75.73 & \cellcolor{yellow!22}69.20
& \cellcolor{yellow!22}82.83 & \cellcolor{yellow!22}72.81
& \cellcolor{red!18}82.12 & \cellcolor{red!18}73.20
& \cellcolor{yellow!22}66.61 & 58.96 \\

\rowcolor{gray!20}
12 cams
& \cellcolor{yellow!22}78.10 & \cellcolor{red!18}69.56
& 80.71 & 70.55
& 80.16 & 70.01
& 66.39 & \cellcolor{yellow!22}60.96 \\

\hline\bottomrule
\end{tabular}
}
\vspace{-3mm}
\end{table}

%% file: table/ablation_vlm.tex
\begin{table}[t]
\centering
\caption{Ablation study on various Vision-Language Models (VLMs). We evaluate the impact of different VLM backbones on our framework's reasoning capabilities across four query categories. Cells highlighted in \textbf{red} and \textbf{yellow} denote the best and second-best performance within each query category.}
\vspace{-3mm}
\label{tab:ablation_vlm}

\setlength{\tabcolsep}{0pt} 
{
    
    \begin{tabular}{lcccccccc}
    \toprule\hline
    \multirow{2.5}{*}{VLM} & \multicolumn{2}{c}{\textbf{Attr. Query}} & \multicolumn{2}{c}{\textbf{Act. Query}} & \multicolumn{2}{c}{\textbf{Spat. Query}} & \multicolumn{2}{c}{\textbf{Inter. Query}} \\
    \cmidrule(lr){2-3} \cmidrule(lr){4-5} \cmidrule(lr){6-7} \cmidrule(lr){8-9}
     & mAcc & mIoU & mAcc & mIoU & mAcc & mIoU & mAcc & mIoU \\
    \midrule

    \rowcolor{gray!20}
    w/o VLM             & 75.49          & 65.03          & 73.98          & 62.87          & 79.15          & 70.30          & 60.22          & 51.28          \\
    Gemini3-Pro         & 76.34          & 66.92          & \cellcolor{yellow!22}83.29          & \cellcolor{yellow!22}73.92          & \cellcolor{red!18}81.52 & 72.18          & \cellcolor{red!18}70.02 & \cellcolor{yellow!22}62.08          \\
    \rowcolor{gray!20}
    GPT5.2              & \cellcolor{red!18}77.04 & \cellcolor{red!18}67.47 & 82.26          & 73.51          & 79.19          & \cellcolor{red!18}73.29 & 69.13          & 61.36          \\
    Qwen2.5-VL (7B)     & 76.16          & 66.71          & 82.96          & 72.94          & 80.21          & 72.42          & 69.39          & 61.94          \\
    
    \midrule

    \rowcolor{gray!20}
    Qwen3-VL (8B) & \cellcolor{yellow!22}76.72          & \cellcolor{yellow!22}67.11          & \cellcolor{red!18}83.66 & \cellcolor{red!18}75.14 & \cellcolor{yellow!22}81.19          & \cellcolor{yellow!22}72.97          & \cellcolor{yellow!22}69.83          & \cellcolor{red!18}63.73 \\
    \hline\bottomrule
    \end{tabular}
}
\vspace{-3mm}
\end{table}

%% file: table/quantitative_all_scene_scale.tex
\begin{table}[t]
\centering
\caption{Average quantitative results on Panoptic-L4D across different spatial scales. We report mAcc(\%) and mIoU(\%) for four query types.}
\label{tab:avg_scene_results}
\vspace{-3mm}
\small
\setlength{\tabcolsep}{2.0pt}

\begin{tabular}{lcccccccc}
\toprule\hline
\multirow{2}{*}{Scale} &
\multicolumn{2}{c}{\textbf{Attr.}} &
\multicolumn{2}{c}{\textbf{Act.}} &
\multicolumn{2}{c}{\textbf{Spat.}} &
\multicolumn{2}{c}{\textbf{Inter.}} \\
\cmidrule(lr){2-3} \cmidrule(lr){4-5} \cmidrule(lr){6-7} \cmidrule(lr){8-9}
& mAcc & mIoU & mAcc & mIoU & mAcc & mIoU & mAcc & mIoU \\
\midrule

Outdoor
& 82.08 & 72.15
& 82.76 & 73.46
& 87.65 & 79.26
& 70.43 & 64.92 \\

\rowcolor{gray!20}
Tabletop
& 68.06 & 59.96
& 77.62 & 68.00
& 84.44 & 72.92
& 75.00 & 62.59 \\

Room
& 76.00 & 68.75
& 90.48 & 80.58
& 71.43 & 64.94
& 73.91 & 66.00 \\

\hline\bottomrule
\end{tabular}
\vspace{-3mm}
\end{table}

%% file: sec/6_conclusion.tex
\section{Conclusion}
\label{sec:conclusion}
We presented PanopticQuery, a framework for language-based understanding in dynamic 4D scenes that separates geometric reconstruction from semantic reasoning. By performing interpretation at query time, our method supports complex queries involving interactions, actions, and spatial relationships that static, embedding-based approaches cannot handle.
A key technical contribution is our multi-view semantic consensus algorithm, which lifts noisy 2D predictions into consistent 4D groundings. We also introduced Panoptic-L4D, the first benchmark targeting contextual and relational reasoning in 4D scenes, where our approach achieves state-of-the-art results across diverse query types.

While effective for short- to medium-term reasoning, our method faces challenges in long-duration scenes where tracking and memory become critical. Future work will explore more efficient, feed-forward architectures and incorporate temporal memory to better scale reasoning over extended timeframes.

%% file: ref.bib
@article{mildenhall2021nerf,
  title={Nerf: Representing scenes as neural radiance fields for view synthesis},
  author={Mildenhall, Ben and Srinivasan, Pratul P and Tancik, Matthew and Barron, Jonathan T and Ramamoorthi, Ravi and Ng, Ren},
  journal={Communications of the ACM},
  volume={65},
  number={1},
  pages={99--106},
  year={2021},
  publisher={ACM New York, NY, USA}
}

@article{li2025langsplatv2,
  title={Langsplatv2: High-dimensional 3d language gaussian splatting with 450+ fps},
  author={Li, Wanhua and Zhao, Yujie and Qin, Minghan and Liu, Yang and Cai, Yuanhao and Gan, Chuang and Pfister, Hanspeter},
  journal={arXiv preprint arXiv:2507.07136},
  year={2025}
}

@article{he2025refersplat,
  title={ReferSplat: Referring segmentation in 3d gaussian splatting},
  author={He, Shuting and Jie, Guangquan and Wang, Changshuo and Zhou, Yun and Hu, Shuming and Li, Guanbin and Ding, Henghui},
  journal={arXiv preprint arXiv:2508.08252},
  year={2025}
}

@inproceedings{ye2024gaussian,
  title={Gaussian grouping: Segment and edit anything in 3d scenes},
  author={Ye, Mingqiao and Danelljan, Martin and Yu, Fisher and Ke, Lei},
  booktitle={European conference on computer vision},
  pages={162--179},
  year={2024},
  organization={Springer}
}

@article{ji2024segment,
  title={Segment any 4d gaussians},
  author={Ji, Shengxiang and Wu, Guanjun and Fang, Jiemin and Cen, Jiazhong and Yi, Taoran and Liu, Wenyu and Tian, Qi and Wang, Xinggang},
  journal={arXiv preprint arXiv:2407.04504},
  year={2024}
}

@inproceedings{cuttano2025samwise,
  title={Samwise: Infusing wisdom in sam2 for text-driven video segmentation},
  author={Cuttano, Claudia and Trivigno, Gabriele and Rosi, Gabriele and Masone, Carlo and Averta, Giuseppe},
  booktitle={Proceedings of the Computer Vision and Pattern Recognition Conference},
  pages={3395--3405},
  year={2025}
}

@inproceedings{li2025seeground,
  title={Seeground: See and ground for zero-shot open-vocabulary 3d visual grounding},
  author={Li, Rong and Li, Shijie and Kong, Lingdong and Yang, Xulei and Liang, Junwei},
  booktitle={Proceedings of the Computer Vision and Pattern Recognition Conference},
  pages={3707--3717},
  year={2025}
}

@article{zhan2025freeq,
  title={FreeQ-Graph: Free-form Querying with Semantic Consistent Scene Graph for 3D Scene Understanding},
  author={Zhan, Chenlu and Zhang, Yufei and Wang, Gaoang and Wang, Hongwei},
  journal={arXiv preprint arXiv:2506.13629},
  year={2025}
}

@inproceedings{li2022neural,
  title={Neural 3d video synthesis from multi-view video},
  author={Li, Tianye and Slavcheva, Mira and Zollhoefer, Michael and Green, Simon and Lassner, Christoph and Kim, Changil and Schmidt, Tanner and Lovegrove, Steven and Goesele, Michael and Newcombe, Richard and others},
  booktitle={Proceedings of the IEEE/CVF conference on computer vision and pattern recognition},
  pages={5521--5531},
  year={2022}
}

@article{park2021hypernerf,
  title={Hypernerf: A higher-dimensional representation for topologically varying neural radiance fields},
  author={Park, Keunhong and Sinha, Utkarsh and Hedman, Peter and Barron, Jonathan T and Bouaziz, Sofien and Goldman, Dan B and Martin-Brualla, Ricardo and Seitz, Steven M},
  journal={arXiv preprint arXiv:2106.13228},
  year={2021}
}

@inproceedings{park2021nerfies,
  title={Nerfies: Deformable neural radiance fields},
  author={Park, Keunhong and Sinha, Utkarsh and Barron, Jonathan T and Bouaziz, Sofien and Goldman, Dan B and Seitz, Steven M and Martin-Brualla, Ricardo},
  booktitle={Proceedings of the IEEE/CVF international conference on computer vision},
  pages={5865--5874},
  year={2021}
}

@inproceedings{pumarola2021d,
  title={D-nerf: Neural radiance fields for dynamic scenes},
  author={Pumarola, Albert and Corona, Enric and Pons-Moll, Gerard and Moreno-Noguer, Francesc},
  booktitle={Proceedings of the IEEE/CVF conference on computer vision and pattern recognition},
  pages={10318--10327},
  year={2021}
}

@article{song2023nerfplayer,
  title={Nerfplayer: A streamable dynamic scene representation with decomposed neural radiance fields},
  author={Song, Liangchen and Chen, Anpei and Li, Zhong and Chen, Zhang and Chen, Lele and Yuan, Junsong and Xu, Yi and Geiger, Andreas},
  journal={IEEE Transactions on Visualization and Computer Graphics},
  volume={29},
  number={5},
  pages={2732--2742},
  year={2023},
  publisher={IEEE}
}

@inproceedings{xian2021space,
  title={Space-time neural irradiance fields for free-viewpoint video},
  author={Xian, Wenqi and Huang, Jia-Bin and Kopf, Johannes and Kim, Changil},
  booktitle={Proceedings of the IEEE/CVF conference on computer vision and pattern recognition},
  pages={9421--9431},
  year={2021}
}

@inproceedings{fridovich2023k,
  title={K-planes: Explicit radiance fields in space, time, and appearance},
  author={Fridovich-Keil, Sara and Meanti, Giacomo and Warburg, Frederik Rahb{\ae}k and Recht, Benjamin and Kanazawa, Angjoo},
  booktitle={Proceedings of the IEEE/CVF Conference on Computer Vision and Pattern Recognition},
  pages={12479--12488},
  year={2023}
}

@inproceedings{cao2023hexplane,
  title={Hexplane: A fast representation for dynamic scenes},
  author={Cao, Ang and Johnson, Justin},
  booktitle={Proceedings of the IEEE/CVF Conference on Computer Vision and Pattern Recognition},
  pages={130--141},
  year={2023}
}

@article{kerbl20233d,
  title={3D Gaussian Splatting for Real-Time Radiance Field Rendering},
  author={Kerbl, Bernhard and Kopanas, Georgios and Leimkuehler, Thomas and Drettakis, George},
  journal={ACM Transactions on Graphics (TOG)},
  volume={42},
  number={4},
  pages={1--14},
  year={2023},
  publisher={ACM New York, NY, USA}
}

@inproceedings{wu20244d,
  title={4d gaussian splatting for real-time dynamic scene rendering},
  author={Wu, Guanjun and Yi, Taoran and Fang, Jiemin and Xie, Lingxi and Zhang, Xiaopeng and Wei, Wei and Liu, Wenyu and Tian, Qi and Wang, Xinggang},
  booktitle={Proceedings of the IEEE/CVF conference on computer vision and pattern recognition},
  pages={20310--20320},
  year={2024}
}

@inproceedings{yang2024deformable,
  title={Deformable 3d gaussians for high-fidelity monocular dynamic scene reconstruction},
  author={Yang, Ziyi and Gao, Xinyu and Zhou, Wen and Jiao, Shaohui and Zhang, Yuqing and Jin, Xiaogang},
  booktitle={Proceedings of the IEEE/CVF conference on computer vision and pattern recognition},
  pages={20331--20341},
  year={2024}
}

@inproceedings{li2024spacetime,
  title={Spacetime gaussian feature splatting for real-time dynamic view synthesis},
  author={Li, Zhan and Chen, Zhang and Li, Zhong and Xu, Yi},
  booktitle={Proceedings of the IEEE/CVF Conference on Computer Vision and Pattern Recognition},
  pages={8508--8520},
  year={2024}
}

@InProceedings{Song_2025_ICCV,
    author    = {Song, Rui and Liang, Chenwei and Xia, Yan and Zimmer, Walter and Cao, Hu and Caesar, Holger and Festag, Andreas and Knoll, Alois},
    title     = {CoDa-4DGS: Dynamic Gaussian Splatting with Context and Deformation Awareness for Autonomous Driving},
    booktitle = {Proceedings of the IEEE/CVF International Conference on Computer Vision (ICCV)},
    month     = {October},
    year      = {2025},
    pages     = {28031-28041}
}

@inproceedings{
    liu2025modgs,
    title={Mo{DGS}: Dynamic Gaussian Splatting from Casually-captured Monocular Videos with Depth Priors},
    author={Qingming LIU and Yuan Liu and Jiepeng Wang and Xianqiang Lyu and Peng Wang and Wenping Wang and Junhui Hou},
    booktitle={The Thirteenth International Conference on Learning Representations},
    year={2025},
    url={https://openreview.net/forum?id=2prShxdLkX}
}

@inproceedings{kerr2023lerf,
  title={Lerf: Language embedded radiance fields},
  author={Kerr, Justin and Kim, Chung Min and Goldberg, Ken and Kanazawa, Angjoo and Tancik, Matthew},
  booktitle={Proceedings of the IEEE/CVF international conference on computer vision},
  pages={19729--19739},
  year={2023}
}

@inproceedings{qin2024langsplat,
  title={Langsplat: 3d language gaussian splatting},
  author={Qin, Minghan and Li, Wanhua and Zhou, Jiawei and Wang, Haoqian and Pfister, Hanspeter},
  booktitle={Proceedings of the IEEE/CVF Conference on Computer Vision and Pattern Recognition},
  pages={20051--20060},
  year={2024}
}

@inproceedings{labe2024dgd,
  title={Dgd: Dynamic 3d gaussians distillation},
  author={Labe, Isaac and Issachar, Noam and Lang, Itai and Benaim, Sagie},
  booktitle={European Conference on Computer Vision},
  pages={361--378},
  year={2024},
  organization={Springer}
}

@inproceedings{fiebelman20254,
  title={4-LEGS: 4D Language Embedded Gaussian Splatting},
  author={Fiebelman, Gal and Cohen, Tamir and Morgenstern, Ayellet and Hedman, Peter and Averbuch-Elor, Hadar},
  booktitle={Computer Graphics Forum},
  pages={e70085},
  year={2025},
  organization={Wiley Online Library}
}

@inproceedings{li20254d,
  title={4d langsplat: 4d language gaussian splatting via multimodal large language models},
  author={Li, Wanhua and Zhou, Renping and Zhou, Jiawei and Song, Yingwei and Herter, Johannes and Qin, Minghan and Huang, Gao and Pfister, Hanspeter},
  booktitle={Proceedings of the Computer Vision and Pattern Recognition Conference},
  pages={22001--22011},
  year={2025}
}

@inproceedings{kirillov2023segment,
  title={Segment anything},
  author={Kirillov, Alexander and Mintun, Eric and Ravi, Nikhila and Mao, Hanzi and Rolland, Chloe and Gustafson, Laura and Xiao, Tete and Whitehead, Spencer and Berg, Alexander C and Lo, Wan-Yen and others},
  booktitle={Proceedings of the IEEE/CVF international conference on computer vision},
  pages={4015--4026},
  year={2023}
}

@article{2023GPT4VisionSC,
  title={GPT-4V(ision) System Card},
  author={OpenAI},
  year={2023},
  journal={OpenAI},
  url={https://api.semanticscholar.org/CorpusID:263218031}
}

@article{Qwen3-VL,
      title={Qwen3-VL Technical Report}, 
      author={Shuai Bai and Yuxuan Cai and Ruizhe Chen and Keqin Chen and Xionghui Chen and Zesen Cheng and Lianghao Deng and Wei Ding and Chang Gao and Chunjiang Ge and Wenbin Ge and Zhifang Guo and Qidong Huang and Jie Huang and Fei Huang and Binyuan Hui and Shutong Jiang and Zhaohai Li and Mingsheng Li and Mei Li and Kaixin Li and Zicheng Lin and Junyang Lin and Xuejing Liu and Jiawei Liu and Chenglong Liu and Yang Liu and Dayiheng Liu and Shixuan Liu and Dunjie Lu and Ruilin Luo and Chenxu Lv and Rui Men and Lingchen Meng and Xuancheng Ren and Xingzhang Ren and Sibo Song and Yuchong Sun and Jun Tang and Jianhong Tu and Jianqiang Wan and Peng Wang and Pengfei Wang and Qiuyue Wang and Yuxuan Wang and Tianbao Xie and Yiheng Xu and Haiyang Xu and Jin Xu and Zhibo Yang and Mingkun Yang and Jianxin Yang and An Yang and Bowen Yu and Fei Zhang and Hang Zhang and Xi Zhang and Bo Zheng and Humen Zhong and Jingren Zhou and Fan Zhou and Jing Zhou and Yuanzhi Zhu and Ke Zhu},
	  journal={arXiv preprint arXiv:2511.21631},
      year={2025}
}

@article{Qwen2.5-VL,
  title={Qwen2.5-VL Technical Report},
  author={Bai, Shuai and Chen, Keqin and Liu, Xuejing and Wang, Jialin and Ge, Wenbin and Song, Sibo and Dang, Kai and Wang, Peng and Wang, Shijie and Tang, Jun and Zhong, Humen and Zhu, Yuanzhi and Yang, Mingkun and Li, Zhaohai and Wan, Jianqiang and Wang, Pengfei and Ding, Wei and Fu, Zheren and Xu, Yiheng and Ye, Jiabo and Zhang, Xi and Xie, Tianbao and Cheng, Zesen and Zhang, Hang and Yang, Zhibo and Xu, Haiyang and Lin, Junyang},
  journal={arXiv preprint arXiv:2502.13923},
  year={2025}
}

@article{Qwen2-VL,
  title={Qwen2-VL: Enhancing Vision-Language Model's Perception of the World at Any Resolution},
  author={Wang, Peng and Bai, Shuai and Tan, Sinan and Wang, Shijie and Fan, Zhihao and Bai, Jinze and Chen, Keqin and Liu, Xuejing and Wang, Jialin and Ge, Wenbin and Fan, Yang and Dang, Kai and Du, Mengfei and Ren, Xuancheng and Men, Rui and Liu, Dayiheng and Zhou, Chang and Zhou, Jingren and Lin, Junyang},
  journal={arXiv preprint arXiv:2409.12191},
  year={2024}
}

@article{Qwen-VL,
  title={Qwen-VL: A Versatile Vision-Language Model for Understanding, Localization, Text Reading, and Beyond},
  author={Bai, Jinze and Bai, Shuai and Yang, Shusheng and Wang, Shijie and Tan, Sinan and Wang, Peng and Lin, Junyang and Zhou, Chang and Zhou, Jingren},
  journal={arXiv preprint arXiv:2308.12966},
  year={2023}
}

@article{ravi2024sam,
  title={Sam 2: Segment anything in images and videos},
  author={Ravi, Nikhila and Gabeur, Valentin and Hu, Yuan-Ting and Hu, Ronghang and Ryali, Chaitanya and Ma, Tengyu and Khedr, Haitham and R{\"a}dle, Roman and Rolland, Chloe and Gustafson, Laura and others},
  journal={arXiv preprint arXiv:2408.00714},
  year={2024}
}

@inproceedings{sabater2017dataset,
  title={Dataset and pipeline for multi-view light-field video},
  author={Sabater, Neus and Boisson, Guillaume and Vandame, Benoit and Kerbiriou, Paul and Babon, Frederic and Hog, Matthieu and Gendrot, Remy and Langlois, Tristan and Bureller, Olivier and Schubert, Arno and others},
  booktitle={Proceedings of the IEEE conference on computer vision and pattern recognition Workshops},
  pages={30--40},
  year={2017}
}

@inproceedings{zhou2025feature4x,
  title={Feature4x: Bridging any monocular video to 4d agentic ai with versatile gaussian feature fields},
  author={Zhou, Shijie and Ren, Hui and Weng, Yijia and Zhang, Shuwang and Wang, Zhen and Xu, Dejia and Fan, Zhiwen and You, Suya and Wang, Zhangyang and Guibas, Leonidas and others},
  booktitle={Proceedings of the Computer Vision and Pattern Recognition Conference},
  pages={14179--14190},
  year={2025}
}
